\documentclass[acmtog,screen,nonacm]{acmart}

\usepackage{booktabs}

\usepackage[ruled]{algorithm2e} 

\usepackage{times}
\usepackage{graphicx}
\usepackage{multirow}
\usepackage{bm}
\usepackage{xcolor}
\usepackage{cleveref}
\usepackage[normalem]{ulem}

\SetAlFnt{\small}
\SetAlCapFnt{\small}
\SetAlCapNameFnt{\small}
\SetAlCapHSkip{0pt}

\acmJournal{TOG}

\newcommand{\agp}[1]{{\color{black}#1}}
\newcommand{\rz}[1]{{\color{black}#1}}

\begin{document}
\title{
RoSI: Recovering 3D Shape Interiors from Few Articulation Images}

\author{Akshay Gadi Patil}
\affiliation{
 \institution{Simon Fraser University}
 \country{Canada}}

\author{Yiming Qian}
\affiliation{
 \institution{Amazon}
 \country{Canada}}

\author{Shan Yang}
\affiliation{
 \institution{Amazon}
 \country{USA}}

\author{Brian Jackson}
\affiliation{
 \institution{Amazon}
 \country{USA}}
 
\author{Eric Bennett}
\affiliation{
 \institution{Amazon}
 \country{USA}}
 
\author{Hao Zhang}
\affiliation{
 \institution{Simon Fraser University, Amazon}
 \country{Canada}}

\begin{abstract}
The dominant majority of 3D models that appear in gaming, VR/AR, and those we use to train geometric deep learning algorithms are {\em incomplete\/}, since they are modeled as {\em surface\/} meshes and missing their interior structures.
We present a learning framework to {\em recover the shape interiors\/} (RoSI) of existing 3D models with only their exteriors from {\em multi-view\/} and {\em multi-articulation\/} images. Given a set of RGB images that capture a target 3D object in different articulated poses, possibly from only {\em few views\/}, our method infers the interior planes that are observable in the input images. Our neural architecture is trained in a {\em category-agnostic\/} manner and it consists of a {\em motion-aware\/} multi-view analysis phase including pose, depth, and motion estimations, followed by interior plane detection in images and 3D space, and finally multi-view plane fusion. In addition,
our method also predicts part articulations and is able to realize and even extrapolate
the captured motions on the target 3D object.
We evaluate our method by quantitative and qualitative comparisons to baselines and alternative solutions, as well as testing on untrained object categories and real image inputs to assess its generalization capabilities. 
\end{abstract}

%
%

%
%

\maketitle

\section{Introduction}
\label{sec:intro}


The emergence of large-scale 3D shape collections~\cite{ShapeNet,fu20203dfuture,mo2019partnet} has propelled the proliferation of neural 3D processing algorithms in computer graphics and vision. However, in the ensuing applications such as 3D recognition, reconstruction, and generation, the focuses of both the algorithms themselves and the evaluation have predominantly been placed on the (external) {\em shapes\/} and their visual appearance, largely overlooking the shape {\em interiors\/}. This is not surprising since the dominant majority of existing 3D models, e.g., those from ShapeNet~\cite{ShapeNet}, the largest and most frequently adopted training dataset, are only available as {\em surface\/} meshes. Even for 3D models carefully crafted by artists, e.g., from the  Amazon-Berkeley Objects (ABO)~\cite{collins2022abo} dataset, only the exterior surfaces were created.

\begin{figure}
\centering
\includegraphics[width=0.99\linewidth]{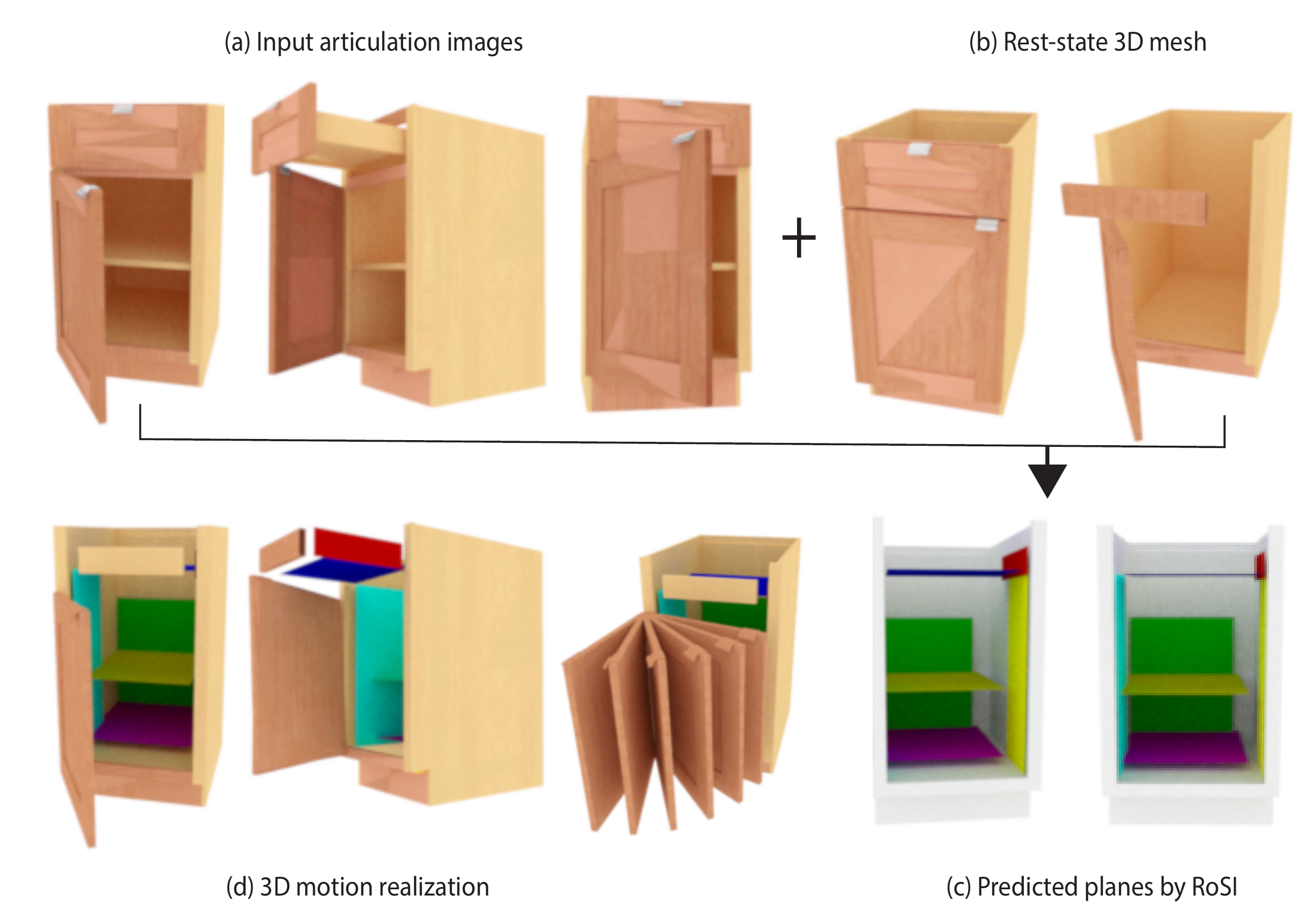}
\caption{RoSI is a learning framework trained to recover the {\em interior\/} of a 3D model from {\em articulation images\/} which capture its various part articulations. We show results produced from three input RGB images (a) for a 3D cabinet provided in its rest (unarticulated) state; see (b)-left. As illustrated in (b)-right by an exploded view, the input 3D model has no interiors. Our method not only infers the interior planes, shown in (c) in solid colors, but can also reproduce, and extrapolate (e.g., door opening), the captured motions on the 3D model (d).}
\label{fig:teaser}
\end{figure}


\begin{figure*}[t]
    \centering
    \includegraphics[width = \linewidth]{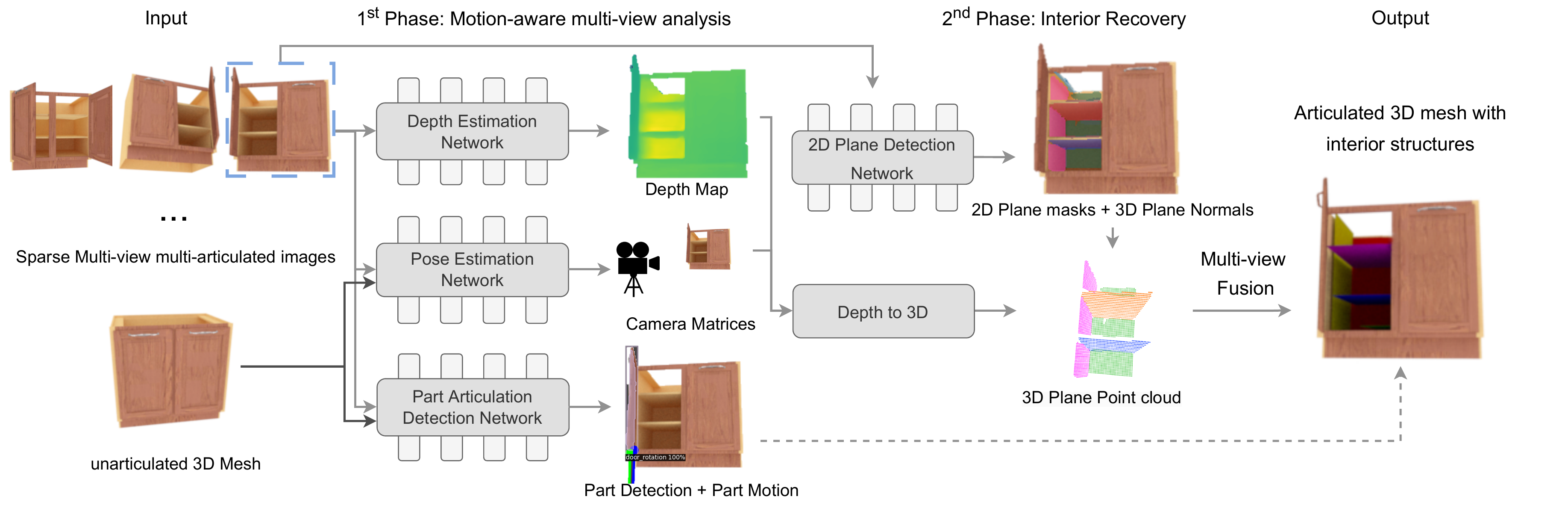}
    \caption{\textbf{Method pipeline.} Separate networks are trained for per-view camera pose estimation, depth estimation, and articulation detection in the first phase. The predicted poses and depths are fed into the second phase for per-view interior plane recovery, followed by multi-view fusion. 
    The rest-state (i.e., unarticulated) 3D model is only used as input to the pose estimation network and optionally, for articulation detection.} 
    \label{fig:pipeline}
\end{figure*}

With only an outside ``shell", a 3D object is {\em incomplete\/} as it is devoid of any of its potentially rich interior structures. Such models cannot adequately support common tasks closely tied to object motion and functionality, e.g., product inspection, physical simulation, and agent-object interactions in robotic and AR/VR.
In particular, as a 3D object is interacted with and undergoes motions, its interiors may be revealed, e.g., the back planes of a drawer or the interior shelving of a cabinet when its door is opened; see Figure~\ref{fig:teaser}.

Our goal in this work is to {\em recover the interior geometries\/} of existing 3D models with only their exteriors from {\em image guidance\/}. To this end, we introduce a method to solve the new problem of {\em multi-articulation\/} recovery of shape interiors (RoSI). Specifically, given a set of RGB images that capture a target 3D object in different articulated poses, possibly from only {\em few views\/}, our method infers the interior planes that are observable from the input images. In our setting, we assume that the target 3D object is given as a ``rest-state'' mesh shell, i.e., a mesh without interior structures or part articulation. In addition to plane recovery, our method also predicts part articulations from the input images and is able to realize the captured motions on the target object in 3D, while exposing the predicted interior planes, as shown in Figure~\ref{fig:teaser} for a cabinet.

It is well known that 3D geometry inference from sparse RGB images is a severely ill-posed problem, even when the outputs are restricted to planes. Recently, several learning-based methods have been proposed for plane detection from images, most notably PlaneRCNN~\cite{liu2019planercnn}. However, a unique challenge in our work comes from the input images, which capture an object undergoing {\em different\/} part articulations, with no constraints on the number of articulated parts in each image. In addition, the input images may exhibit significant diversity when they were captured from varying camera poses, both in terms of viewing distance and angle. These, along with the lack of image overlap due to view sparsity, would significantly compromise conventional multi-view 3D reconstruction approaches built on image correspondence. An additional implied difficulty arises when estimating camera poses, which are unknown but necessary to place the per-view predictions into a common coordinate frame, for plane integration and motion realization.

As outlined in Figure~\ref{fig:pipeline}, our method constitutes a learning framework with two main phases. First, we perform {\em motion-aware multi-view analysis} which takes as input the multi-view, multi-articulation images and the rest-state, unarticulated 3D model, and predicts per-view camera poses, depth maps, and part articulations, using separate neural networks. Among these three tasks, camera pose prediction for the given rest-state 3D model from articulation images is particularly challenging due to the lack of correspondences, mainly over regions of articulated parts. To address this new challenge, we design a motion-aware neural architecture which simultaneously detects unarticulated image pixels and leverages them for feature matching, as detailed in Section~\ref{sec:pose_est}.
%
%
For articulation detection (Section~\ref{sec:mot_det}), we build upon OPD \cite{jiang2022opd} and use the detected parts and associated motion parameters to transfer motions to 3D.

In the second phase, we recover shape interiors as planes in 3D (Section~\ref{sec:interior_recovery}). Our network detects per-view 2D plane masks corresponding to the interior regions and infers their associated 3D plane normals in the camera coordinate frame. The interior planes in 3D space are then recovered based on the estimated camera pose, depths, per-view, and then merged in the common world frame via thresholding.
Our neural model is built on PlaneRCNN~\cite{liu2019planercnn}, which was originally proposed for single-view plane detection in the camera frame. Although we process each articulation image independently, the key difference in our problem setting lies in projecting the view-recovered planes to a common coordinate frame for plausible interior shape recovery, which hinges critically on the the pose estimation module. Putting these together, our contribution also includes proposing a system, which is the first one catered towards interior shape recovery from multi-articulation images.

All the modules in our method are trained using the PartNet-Mobility dataset~\cite{xiang2020sapien} which provides $\approx$2K 3D models with interior structures; the training is {\em category-agnostic\/}. Models are rendered using Mitsuba renderer~\cite{jakob2022mitsuba3}, with an average of 19 view images per model (eventually, only 3 are considered), by random part articulations. Quantitative and qualitative evaluations show that our method achieves promising results by comparing against baselines and alternative solutions. We also perform tests on untrained object categories and on \rz{real images from the ABO dataset to asses the \rz{generalization capability} of our approach.}

\section{Related work}
\label{sec:related}

In this section, we cover related works on 3D reconstruction, especially plane inference from images, neural 3D representation and reconstruction focusing on articulation, articulation detection from images, and pose estimation. We also discuss shape completion as a possible alternative for recovering 3D shape interiors.

\vspace{-5pt}

\paragraph{3D reconstruction}
3D reconstruction is one of the most studied problems in computer graphics~\cite{berger2017survey}.
%
In typical problem settings, a target 3D object is captured from one or more views in the form of RGB images, depth values, or point clouds, from which the 3D shape is to be recovered.
While most 3D objects we encounter every day, e.g., common household items including furniture and appliances, exhibit various motions and articulations to perform their functions, conventional reconstruction methods assume that the target object maintains the {\em same\/} static and unarticulated pose in all views. This assumption is essential to a key step of the reconstruction, namely, multi-view registration or feature matching. At best, the resulting methods can only fully recover the object exteriors.

\vspace{-5pt}

\paragraph{Recovering object interiors}
To our knowledge, {\em proactive scanning\/}~\cite{yan2014proactive} is the only classical 3D reconstruction approach geared towards hidden object interiors. In this setting, human users actively adjust the target scene, e.g., opening a cabinet door or drawer, while continuously scanning it to reveal occluded regions. Despite its potential, this 3D acquisition paradigm introduces many technical challenges involving scan registration, human occlusion, as well as motion tracking and recognition.
%
In clear contrast, our approach does {\em not\/} scan the object interiors; it relies on supervised learning to recover shape interiors from articulation images, while limited to what is visible. The rest-state 3D model is only available as a shell and its part articulations must be inferred from the images as well.

Shape completion is a possible alternative for synthesizing interior geometries, when the input consists of an exterior shell. While there have been many neural shape completion methods, mostly operating on point clouds~\cite{yuan2018pcn, pan2021variational, huang2020pf} and with recent attempts utilizing transformers~\cite{zhou2022seedformer, xiang2022snowflake, yan2022shapeformer, yu2021pointr}, we are not aware of any work that specifically targets the recovery of shape interiors.

%

\vspace{-5pt}

\paragraph{Neural reconstruction with articulation}
Some recent works learn {\em category-specific\/} neural shape and appearance representations for articulated 3D shapes, e.g, NASAM~\cite{wei2022self} and A-SDF~\cite{mu2021sdf}, where multi-view inference has been considered as an application. Most closely related to our work is StrobeNet~\cite{zhang2021strobenet}, which reconstructs articulated 3D objects from multi-view and multi-articulation images. However, like the other works, StrobeNet also learns a category-level neural reconstruction model, i.e., it is trained per category and not dedicated to recover 3D shape interiors. In contrast, RoSI targets the specific problem of interior plane recovery and is trained in a {\em category-agnostic} manner.

\vspace{-5pt}

\paragraph{Plane reconstruction from images}
Early works \cite{furukawa2009manhattan, gallup2010piecewise, silberman2012indoor, sinha2009piecewise, zebedin2008fusion} on planar surface reconstruction assign plane proposals over point clouds via global inference. In recent years, neural piece-wise planar reconstruction from a \emph{single} image has gained quite a lot of interest, resulting in end-to-end architectures as proposed in \cite{liu2018planenet, liu2019planercnn, yu2019single, tan2021planetr}. All these methods take one RGB image as input and output a depth image, plane segmentation masks and corresponding plane normals. 3D planar reconstruction is then made possible using the predicted depth map and camera intrinsic matrix. These methods all work on single-view images of indoor scenes, with no articulated objects. Our work deals with multi-view images of articulated 3D objects, where we detect interior planes from every view-image of the articulated model. And since the focus is to recover the interior geometry, obtaining a quality 2D reconstruction of interior depth maps is critical to the end solution. As such, we predict the depth maps separately using a transformer model \cite{ranftl2021vision}, and build upon recent architectures \cite{liu2019planercnn, qian2022understanding} to output interior planes from each articulated view image. 

\vspace{-5pt}

\paragraph{6DoF camera pose estimation} 
Standard pose estimation aims to predict the orientation and translation of an imaged object relative to a canonical CAD model. 
%
%
%
%
The majority of prior methods assume that the image and the canonical model exhibit the same articulation state. To mitigate the performance drop due to object motions, various approaches have been dedicated to handle articulated objects, e.g., pose voting with random forest~\cite{michel2015pose}, articulation graph formulation with Markov Random Field~\cite{desingh2019factored,pavlasek2020parts} from depth maps, pose transformation from optical flow on video inputs~\cite{liu2020nothing}. A number of works focus on category-level pose estimation by extending the Normalized Object Coordinate Space to normalize articulations into canonical motion states~\cite{li2020category,XueLXFL21OMAD,weng2021captra,liu2022toward}. The problem is also actively studied in Robotics where robotic arms are used for interaction~\cite{hausman2015active,katz2008manipulating}. In this paper, we adopt a two-stage approach while proposing a {\em motion-aware\/} and {\em category-agnostic\/} neural architecture for simultaneous correspondence matching and articulation segmentation.

\vspace{-5pt}

\paragraph{Articulation detection from images}
Akin to 2D object detection, the detection of articulated parts of furniture models from images aims at predicting the 2D part bounding box and part mask, as well as inferring various motion-related parameters on the articulated part such as motion type, motion axis, motion origin and motion magnitude. This finds utility in being able to transfer predicted motions from one articulated product image to many similar-looking static 3D models, thus helping create articulated 3D assets.
To achieve this, a natural solution is to extend standard 2D object detection architectures, such as MaskRCNN \cite{he2017mask}. Recent works from \cite{jiang2022opd} and \cite{qian2022understanding} build upon the MaskRCNN architecture, where the former focuses on detecting \emph{openable} parts in articulated images and the latter doing so on articulation videos. Note that they do not focus on predicting motion magnitude. In our work, we build upon the work of \cite{jiang2022opd} and detect all four motion parameters, in addition to 2D part bounding boxes and segmentation masks. We use these estimated 2D articulation parameters in our motion recovery step, as described in Section~\ref{mot_in_3D}.

\section{Overview}
\label{sec:overview}

Figure \ref{fig:pipeline} shows an overview of our system.\\
\textit{System input}: The input to our system is a set of $N$ view-images, $\hat{I}$=$\{I_{1}, I_{2}, ..., I_{N}\}$, of an articulated 3D model, rendered from varying camera poses (both distance of the 3D model from the camera and the viewing angle). As well, each view-image captures varying articulations of the 3D model, in terms of the number of articulated parts and the magnitude of articulation per part. In addition, our system takes as input, the associated part-segmented (unlabeled), interior-devoid 3D model, $S$, in the rest state.\\
\textit{System output}: With the above inputs, our goal is to: (a) recover interior structures revealed by images in $\hat{I}$, and (b) reproduce each input image in $\hat{I}$ in their associated articulations, in 3D.

%
As such, our problem tackles multiple tasks: learning to estimate pose of the 3D model $S$ per view, learning to estimate articulations from a single image, and learning to reconstruct the interior geometry. In our work, we represent the interior geometry using plane as representation. We describe each task below.





\section{Pose estimation from images}
\label{sec:pose_est}

\begin{figure}
    \centering
    \includegraphics[width = \linewidth]{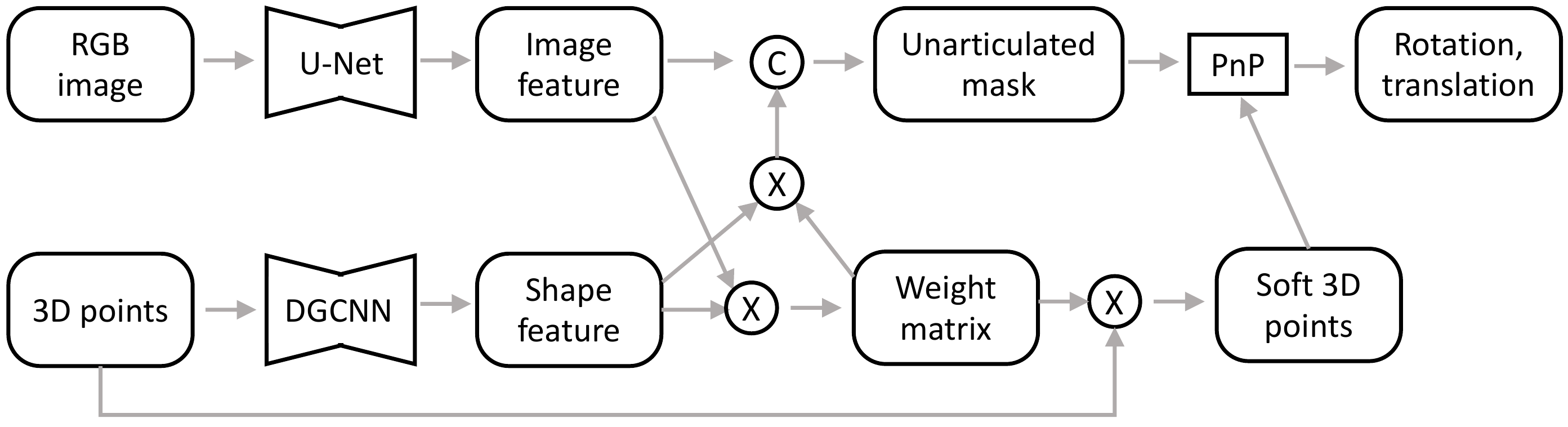}
    \caption{Our motion-aware architecture for pose estimation. Softmax and sigmoid functions are omitted in the figure for brevity.}
    \label{fig:pose_net}
\end{figure}

We propose a two-stage framework to estimate camera pose from articulation images as shown in Figure \ref{fig:pose_net}. Given multi-articulation RGB images and a rest-state 3D model, we first establish a pixel-to-point correspondence between every image and the 3D model defined in a global frame. To this end, a motion-aware neural architecture is proposed for simultaneously identifying unarticulated pixels (i.e., pixels belonging to unarticulated parts of the 3D model) and leveraging them for feature matching. Specifically, the RGB image ($256\times 256 \times 3$) is passed into a U-Net~\cite{ronneberger2015u} resulting in an image embedding of $256\times 256\times 32$ dimensions. We also sample $2048$ points from the rest-state 3D mesh, 
which are processed using DGCNN~\cite{wang2019dynamic}, resulting in a $2048\times 32$ shape embedding. Instead of finding a hard mapping for each pixel, we generate a corresponding 3D point in a differentiable way. That is, we multiply the flattened image embedding with the shape embedding and apply a softmax function, resulting in a $256^2\times2048$ weight matrix. We then multiply it with the 3D point set and obtain the final ``soft" points. Notice that such a soft point generation is also employed in point cloud registration~\cite{wang2019deep}.

Since object exhibits articulation in the captured image, feature matching is valid only at unarticulated pixels. To identify them, we multiply the weight matrix with the shape embedding, concatenate the multiplication result with the image embedding, and obtain a $256\times 256\times 64$ tensor. We pass this tensor to a 2-layer CNN followed by a sigmoid function to output a binary segmentation mask. Our network is trained with L1 loss for the soft point generation branch and with the binary cross entropy loss for the segmentation branch. Finally, we compute the pose parameters using the unarticulated pixels with the Perspective-n-Point (PnP) algorithm.

\begin{figure}[!t]
    \centering
    \includegraphics[width = \linewidth]{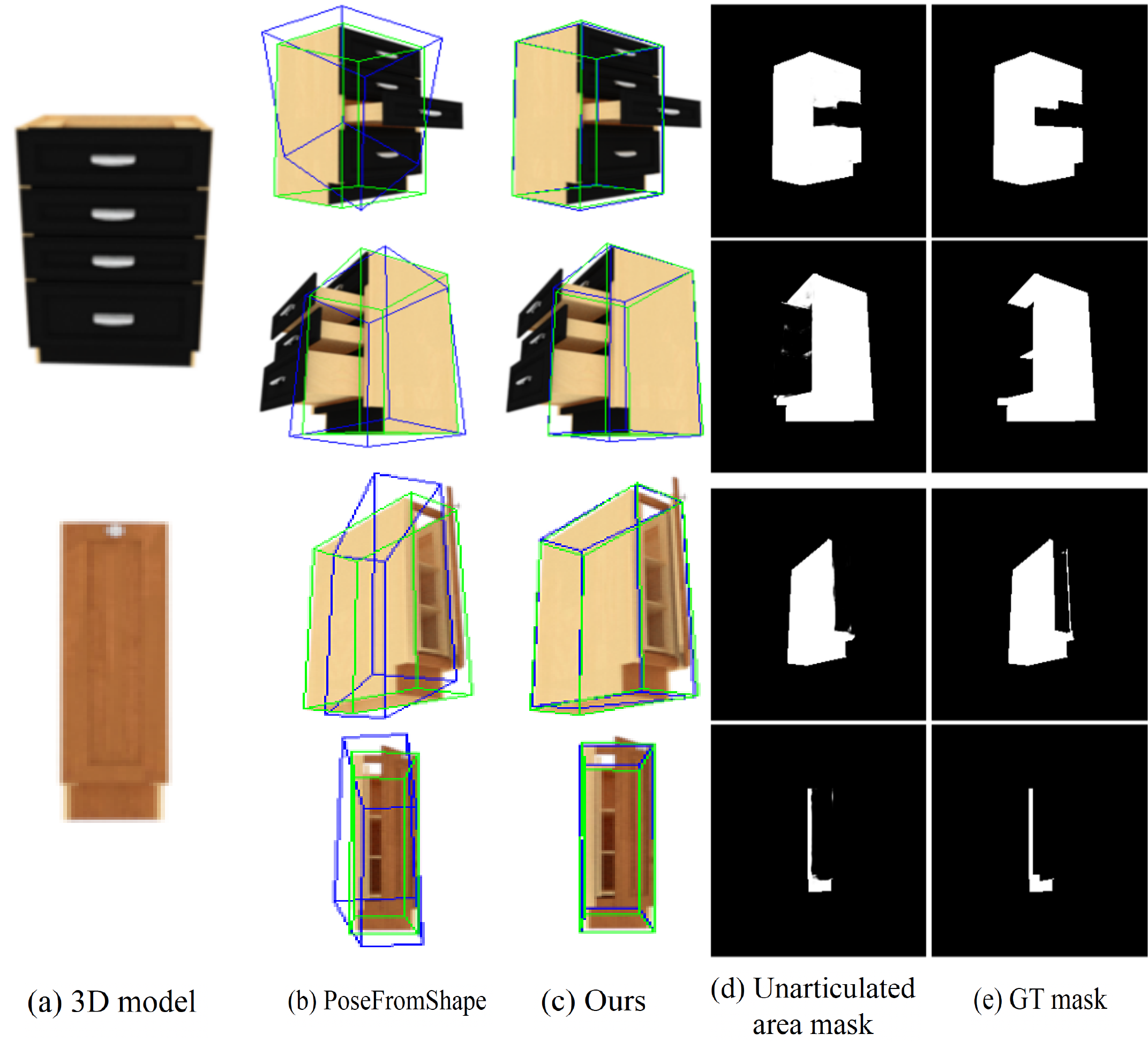}
    \caption{Qualitative results on pose estimation. We show results from two viewpoints in each example. Left to right: rest-state 3D model, 3D bounding box overlaid on image using PoseFromShape, 3D bounding box using our method, our unarticulated area mask, and the GT mask. Green and blue boxes denote the GT and the prediction, respectively. Images in column (b) and (c) are enlarged for visualization purpose.}
    \label{fig:pose_est}
\end{figure}
\section{Estimating articulations from images}
\label{sec:mot_det}

Given a single RGB(D) image $I_{j}$ $\in$ $\{I_{1}, I_{2}, ..., I_{N}\}$, we aim to detect, in 2D, all the articulated parts $\{x_{1}, x_{2}, .., x_{k}\}$ of the furniture model, and also predict, in 3D, the associated motion parameters $\{m_{1}, m_{2}, .., m_{k}\}$. The output, thus, is a set of articulated parts and their associated motion parameters. Each part $x_{i}$ is represented by the set $\{label_{i}, bbox_{i}, mask_{i}\}$, where $label_{i}$ represents the part label (one of the three labels -- lid, drawer, door), $bbox_{i}$ represents the 2D bounding box of the detected part, and $mask_{i}$ represents the 2D segmentation mask. Each motion parameter $m_{i}$ is represented by the set $\{t_{i}, o_{i}, a_{i}, v_{i}\}$, where $t_{i}$ represents the motion type (revolute/prismatic), $o_{i}$ represents the motion origin (3$\times$1 vector), $a_{i}$ represents the motion axis (3$\times$1 vector) and $v_{i}$ represents the motion value/magnitude (scalar). For prismatic motions (eg. pulling out a drawer), $m_{i}$ does not contain $o_{i}$ since motion origin is meaningful only for revolute motions (eg. rotating a door around a hinge).\\

\begin{figure}[!t]
    \centering
    \includegraphics[width = \linewidth]{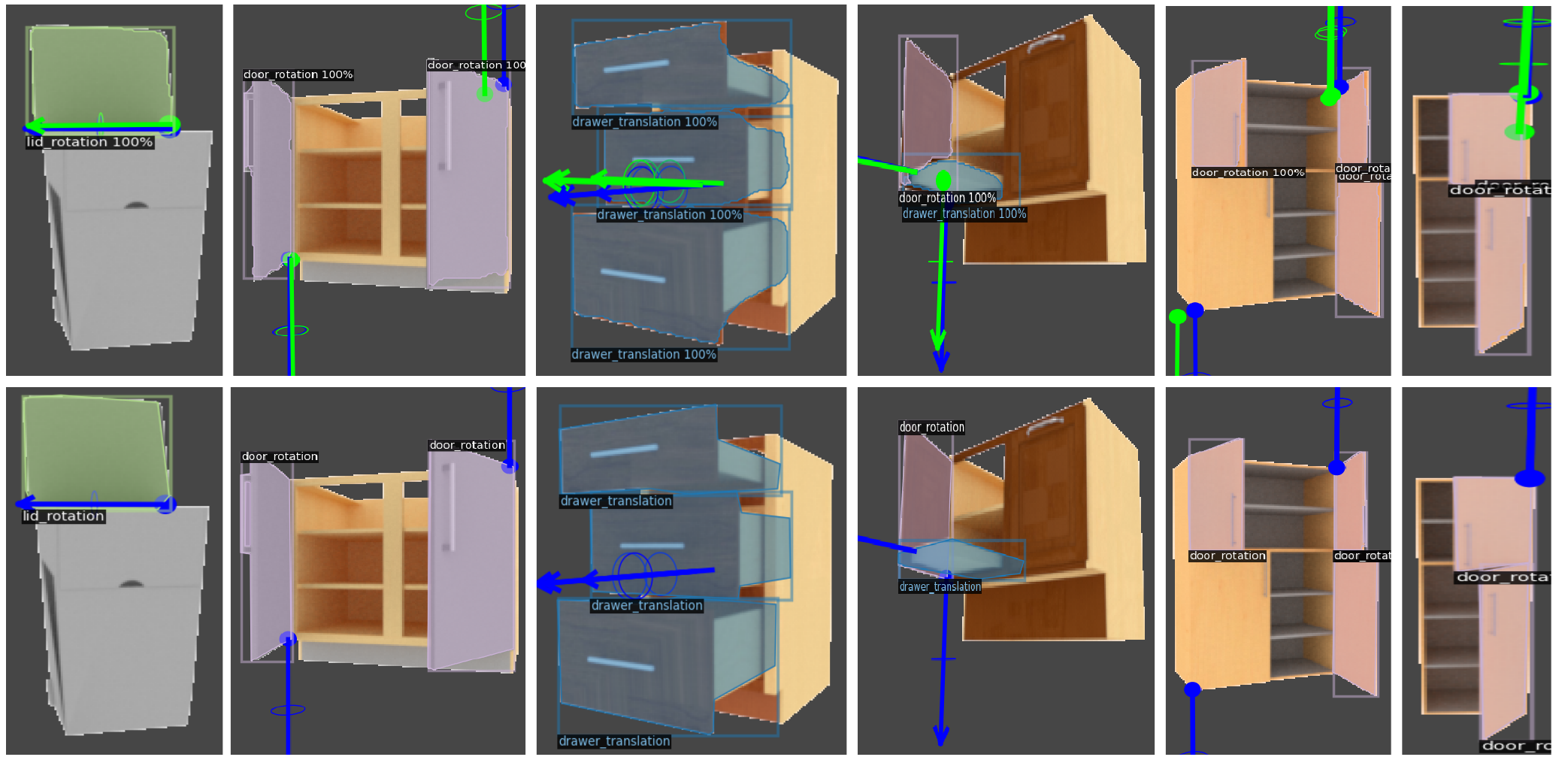}
    \caption{Qualitative results on articulation detection from a single image, with the top row showing network predictions and the bottom row shows the ground truth (GT). The arrows indicate the axis of motion and the small circles at the arrow indicate motion origin. Blue color is for GT motions and green color indicates that the motion predictions are in the local neighborhood of the ground truth.}
    \label{fig:art_detection}
\end{figure}

This problem draws parallels to the task of 2D object detection and instance segmentation, where each articulated part can now be thought of as an object instance. To this end, we build upon the Mask-RCNN architecture \cite{he2017mask} which uses a Faster-RCNN style network for proposing bounding boxes on articulated parts, which are then processed using additional heads to output the instance label, 2D bounding box and 2D segmentation mask. Since we predict motion parameters for every detected articulated part, we add four independent branches at the output of RoIAlign layer to predict these motion parameters $m_{i}$ = $\{t_{i}, o_{i}, a_{i}, v_{i}\}$, in the camera coordinate frame. 
This network is trained in a supervised manner, and the loss function used to optimize the network weights is given by:
\begin{equation}
\label{eqn:mot_det_loss_func}
    l_{total} = (l_{label} + l_{bbox} + l_{mask}) \\ 
    + (l_{mtype} + l_{maxis} + l_{morigin} + l_{mmag})
\end{equation}
where the first three loss terms are the same as in Mask-RCNN \cite{he2017mask}, and the next four loss terms correspond to the four motion prediction heads. Specifically, $l_{label}$ is the negative log likelihood loss for part label, $l_{bbox}$ is the $L_{1}$ regression loss on the 4$\times$1 vector representing the 2D part bounding box, $l_{mask}$ is the average per-pixel sigmoid loss on the binary masks, $l_{mtype}$ is the binary cross-entropy loss for motion type, $l_{maxis}$, $l_{morigin}$ are the $L_{1}$ loss on the motion axis and motion origin, respectively, and $l_{mmag}$ is the MSE loss on the motion value. We do not make use of the predicted part labels (lid/drawer/door) in any of our modules (Figure \ref{fig:pipeline}). A loss against part labels is necessary for learning to localize articulated parts.


\subsection{Motion realization in 3D}
\label{mot_in_3D}
In order to transfer the predicted motions to 3D, we need to be able correspond an exterior part in the rest state segmented 3D model to a detected instance of part articulation. To do this, we predict the depth map corresponding to the input RGB image, as described in Section \ref{sec:interior_recovery}. 
Using the predicted camera pose and the articulation mask, we project the 2D depth image to 3D coordinate space, and reverse/undo the motion on the articulated part using the predicted motion parameters. We then find a one-way Chamfer Distance between each exterior mesh in the rest state segmented 3D model to the articulation-reversed 3D part pointcloud. The exterior mesh that gives us the least distance is then transformed from its rest state using the predicted motion parameters to achieve motion recovery, as shown in Figure \ref{fig:teaser} and \ref{fig:gallery}.
\section{Recovering interior geometry}
\label{sec:interior_recovery}

\subsection{Per-view plane detection}
To recover interior geometry from the set of input images, we first detect, for every input image, all 2D plane masks corresponding to the interior regions and their associated 3D plane normals in the camera co-ordinate frame. Formally, given a single RGB image $I_{j}$ $\in$ $\{I_{1}, I_{2}, ..., I_{N}\}$, our goal is to predict, in 2D, all interior planes $\{p_{1}, p_{2}, .., p_{r}\}$, and also predict, in 3D, the associated plane parameters $\{\pi_{1}, \pi_{2}, .., \pi_{n}\}$. The output, thus, is a set of 2D interior plane masks and the associated planes in 3D.
Each detected plane in 2D, $p_{i}$, is represented by the set $\{label_{i}, bbox_{i}, mask_{i}\}$, where $label_{j}$ represents the mask label (one of the two classes -- planar/non-planar region), $bbox_{j}$ represents the 2D bounding box of the detected interior plane, and $mask_{j}$ represents the 2D segmentation mask of the plane. Each plane in 3D, $\pi_{i}$, corresponding to $p_{i}$, is parameterized by $[\boldsymbol{n_{i}}, o_{i}]$, with $\boldsymbol{n_{i}}$ being the plane normal, representing the coefficients $[a,b,c]$, and $o_{i}$ being the plane offset, such that $\pi^{T}_{i}[x,y,z,1]=0$.\\
%
%
\begin{figure}[!t]
    \centering
    \includegraphics[width = \linewidth]{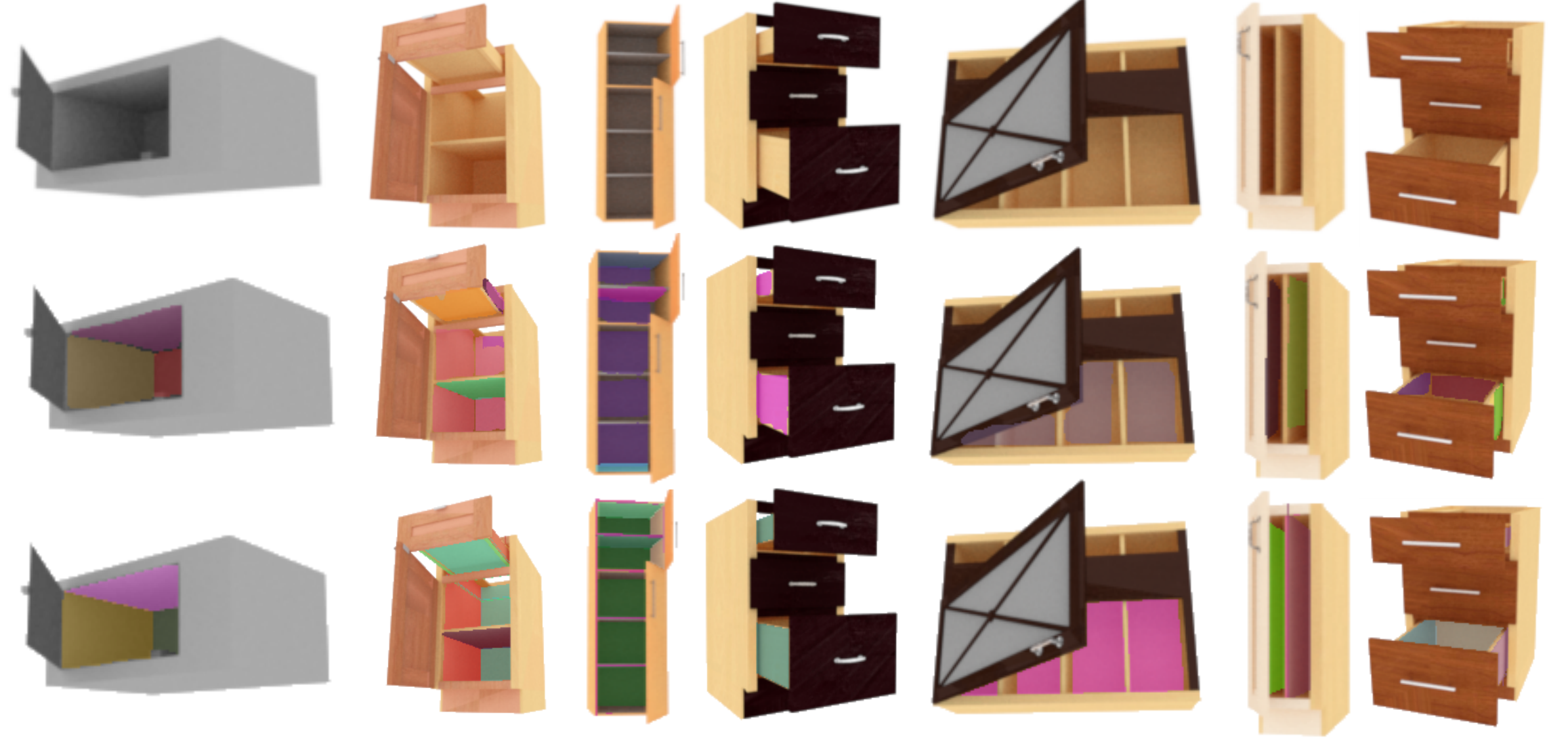}
    \caption{Results on interior plane detection where each predicted plane is visualized using a different color. The top row shows the input image, middle shows predictions, and bottom row is for the ground truth.}
    \label{fig:plane_det}
\end{figure}

Existing works \cite{liu2019planercnn, jiang2020peek,jin2021planar} have shown the adeptness of R-CNN style networks in predicting plane representations. We follow suite. Similar to estimating articulations from images (Section \ref{sec:mot_det}), we employ Mask-RCNN architecture for the above task. So now, each planar region corresponding to the object's interior is treated as an object instance (see Section \ref{subsec:dataset} on how an object region is determined as interior). The plane prediction network then detects such instances and estimates their segmentation masks and plane normals in the camera coordinate frame. The missing plane parameter, the plane offset $o_{i}$, corresponding to $\pi_{i}$, is obtained from the predicted depth maps, as explained below.

\paragraph{Depth estimation.} To correctly recover the planes in 3D, it is essential to have a high-quality depth image corresponding to the input RGB image that would enable a not-so-noisy projection of the detected 2D plane masks to the 3D coordinate space. To this end, we fine-tune a transformer-based monocular depth estimation network \cite{ranftl2021vision} on our data, whose performance was observed to be better, particularly on the interior regions, compared to when the depth map was jointly predicted with the part masks and motion parameters.

\paragraph{Plane offset estimation.}
Recall that the plane normals are estimated in the camera coordinate frame. As such, we use the 3$\times$3 camera intrinsic matrix, $K$, the predicted depth image and the estimated plane normals to obtain the offset as:

\begin{equation}
\label{eqn:offset_cal}
d = \frac{\sum_{i} m_{i}(\boldsymbol{n^{T}_{i}}(z_{i}K^{-1}\boldsymbol{x_{i}}))}{\sum_{i} m_{i}},
\end{equation}

where $\boldsymbol{x_{i}}$ is the $i_{th}$ pixel coordinate in the homogeneous representation, $z_{i}$ is its predicted depth value, and $m_{i}$ is a binary indicator for $\boldsymbol{x_{i}}$ belonging to the 2D mask of plane $p_{i}$. 

\subsection{Multi-view plane fusion}
\label{mv-fusion}
In this step, we combine predicted planes from multiple views by projecting from their camera coordinate frame to the world coordinate frame using predicted camera poses obtained from Section \ref{sec:pose_est}. Depending on the viewing angle, some number of planes coming from different view-images may overlap partially or almost completely, or maybe offset by small amounts. This indicates to the fact that they represent the same interior region. But since each image is processed independently, the relationship between them, i.e., inter-plane relation, is not encoded in any way to ``stitch" them together.

As such, we merge such planes, in a pairwise manner, into a single plane if the plane normal difference between them is less than 15 degrees and if the smaller plane can be subsumed by the larger plane within a given threshold distance (which is 0.3 units in our case). Points from the smaller plane are projected onto the bigger plane if the above conditions are satisfied. Since a 3D shell is made available during testing, it can be used to regularize the position of the planes to fall within its boundaries. We visualize the planes as rectangles in our paper by using bounded values along the two plane axes. 


\section{Results and evaluation}
\label{sec:exp}

\begin{figure*}
\centering
\includegraphics[width=0.99\linewidth]{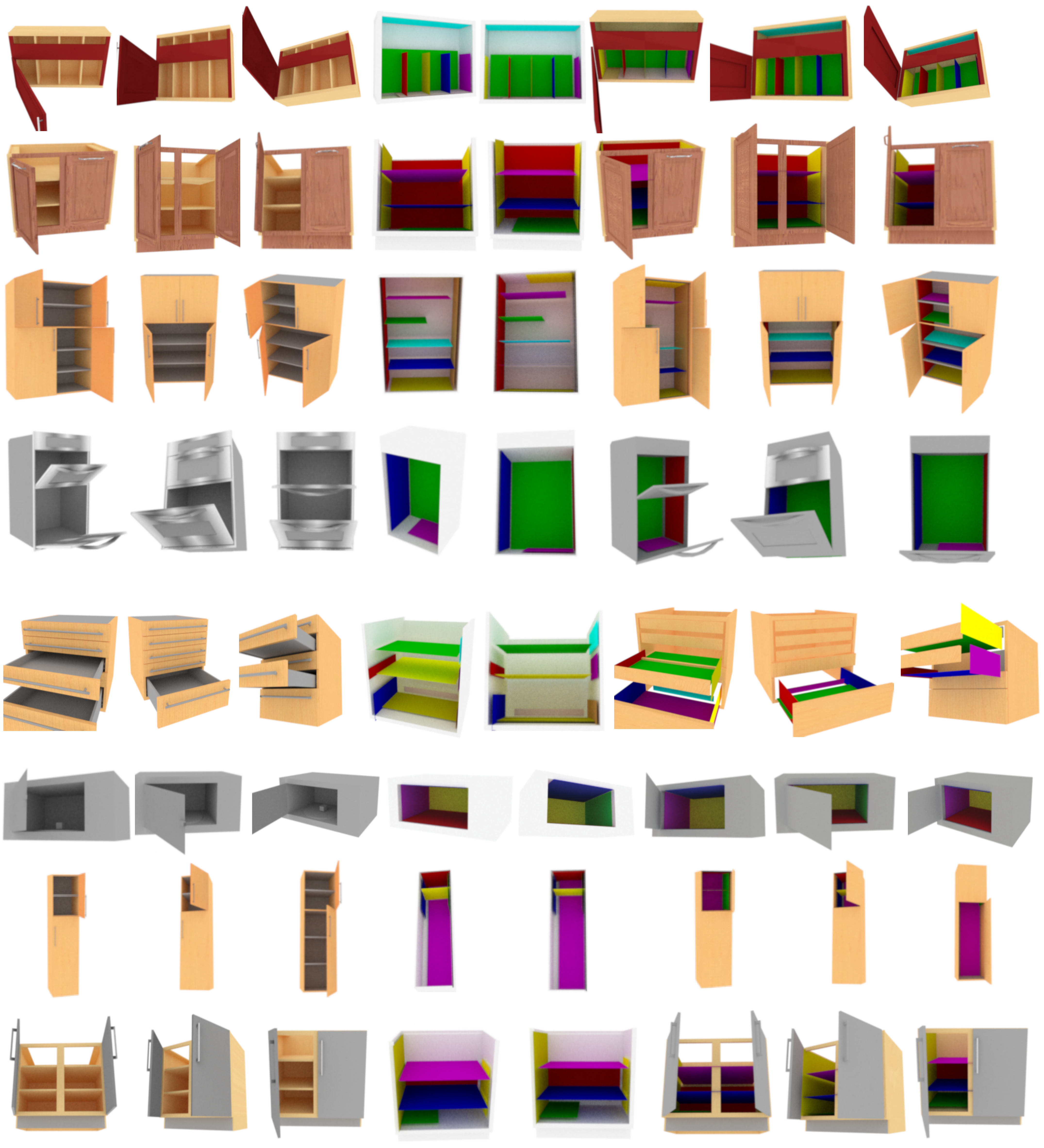}
\caption{Qualitative results on interior recovery from three multi-articulation images, The first three columns show the input, the next two columns show the recovered planes visible in the input images, overlayed on the input 3D shell (semi-transparent) corresponding to the input images. The next three columns show the articulated 3D models with interior structures, rendered from the same viewing angle as the input images. Articulations are realized in 3D based on part detection and motion predictions from individual images.}
\label{fig:gallery}
\end{figure*}

We perform a module-wise evaluation of our system, providing insights on their individual strengths/weaknesses. In addition, we also quantitatively compare our interior recovery module against a baseline. Our pipeline is implemented in PyTorch, on a Tesla V-100 GPU.
Figure \ref{fig:gallery} presents a gallery of results obtained from RoSI.

\begin{table}
    \centering
    \caption{Quantitative evaluations of our motion-aware pose estimation network against PoseFromShape~\cite{Xiao2019PoseFromShape} and a variant of our method without the segmentation branch.} 
    \begin{tabular}{c|c|c}
    ~             & Rotation error ($^{\circ}$) & Translation error \\ \hline
    PoseFromShape & 14.06              & 0.25                 \\
    Ours (w/o segmentation) & 9.53 & 0.13 \\
    Ours          & \textbf{7.61}              & \textbf{0.10}                \\
    \end{tabular}
    \label{tab:pose_result}
\end{table}



\vspace{-5pt}

\paragraph{Evaluation metrics}
To evaluate the pose estimation module, we resort to the standard metrics of angular errors over predicted camera orientation and mean-squared error over translation, compared against the ground truth. 
We also measure the errors in predicted motion parameters, for both, revolute and prismatic motions, separately.
As well, since our system eventually outputs interior planes in 3D, we compare its ability to recover such planes by using the following metrics: (a) measuring geometric similarity via a {\em one-way\/} Chamfer Distance (CD) {\em to the ground truth\/} (GT) planes, (b) mean normal angle difference in degrees, and (c) mean of the plane parameter differences with the ground truth plane.
\begin{table}
    \centering
    \caption{Mean prediction errors on different motion parameters for RGB(D) image inputs from the test set: Raw network predictions vs. post-processing these raw predictions. $R$ refers to Revolute motions and $P$ for Prismatic. Only motion axis and origin are processed, since they are key to plausible motion realization in 3D.}
    \begin{tabular}{c|c|c|c|c|c|c|c}
    Error  & Raw predictions & With post-processing \\ \hline
    ($R$) Axis angle($^{\circ}$) ($\downarrow$) & 16.58 & 14.5 \\
    ($R$) Motion origin ($\downarrow$) & 3.69 & 0.64 \\
    ($P$) Axis angle($^{\circ}$) ($\downarrow$) & 23.47 & 3.16 \\
    ($R$) Magnitude($^{\circ}$) ($\downarrow$) & 28.6 & - \\
    ($P$) Magnitude ($\downarrow$) & 0.165 & -
    
    \end{tabular}
    \label{tab:mot_pred_errors}
\end{table}


%
\subsection{Dataset}
\label{subsec:dataset}
We use the PartNet-Mobility dataset \cite{xiang2020sapien} for all our experiments, which is a part-segmented 3D shape dataset with ground-truth annotation on part motions. We consider 9 categories of objects: Storage, Bin, Fridge, Microwave, Washer, Dishwasher, Oven, Safe and Box. When moveable parts in shapes belonging to these categories are articulated, they reveal planar interior regions, which is the reason such categories were selected to work with (as opposed to categories such as Stapler, Pliers, Scissors, USB etc.). 

Since no annotations for object interiors are available, we identify interior regions by ray-casting. We shoot rays from a source and find faces on the part-segmented 3D object mesh the rays intersect with. An articulated 3D model reveals new intersecting faces that are not otherwise present when using the rest-state 3D model. These new faces are then said to correspond to the object's interior. This setting allows us to obtain ground-truth depth, camera pose, binary articulated part masks (in 2D) and binary interior region masks (in 2D). To obtain ground-truth plane parameters (in 3D) for the interior planes, we use multi-plane RANSAC algorithm \cite{fischler1981random} on the 3D point cloud of 2D interior masks. Images are rendered using the Mitsuba renderer \cite{jakob2022mitsuba3} at $256^{2}$ resolution. We assume a consistent camera intrinsic matrix for all images. More details can be found in the \agp{supplementary}.
In total, after data filtering based on incorrect motion annotations and parts with few faces, we consider a total of 350 shapes, with a 75\%-25\% train-test split.

\subsection{Evaluation on motion predictions}
Motion parameters in 3D allow us to recover motions depicted in images. We therefore evaluate motion predictions from the articulation detection network. We observe that the raw prediction results from the network are in the local neighborhood of the actual solution. As such, we post-process these predictions based on the observation that all models are axis-aligned in the canonical coordinate frame and that axes of motions align with the canonical axes. Do note that the starting solution for such post-processing is the network output. In Table \ref{tab:mot_pred_errors}, we compare the errors from these two methods. Motion origin error is the $L_{2}$ loss between the ground truth and the prediction.


\begin{table}
    \centering
    \caption{We compare to a RANSAC-based baseline for plane fitting on the predicted depth maps for interior plane predictions and recovery, \rz{and to shape completion;} see Section \ref{subsec:eval_int_recovery} for details.}
    \begin{tabular}{c|c|c|c}
    ~                 & Baseline & Ours & Shape completion\\ \hline
    One-way CD ($\downarrow$) &  0.0198          &  \textbf{0.0137}  & 0.151 \\ 
    Normal diff ($^{\circ}$) ($\downarrow$) & 22.421 & \textbf{5.700} & - \\ 
    Param diff ($\downarrow$) & 0.500 & \textbf{0.273} & - \\
    
    \end{tabular}
    \label{tab:baselines}
\end{table}

\subsection{Evaluation on pose estimation}
We compare our approach with another category-agnostic approach PoseFromShape \cite{Xiao2019PoseFromShape}, which directly regresses pose parameters with a neural network. Vanilla PoseFromShape predicts object orientation only, whereas we attach a linear layer to their network to estimate translation and retrain their network on our dataset. Table~\ref{tab:pose_result} shows that our approach outperforms PoseFromShape in both rotation error and translation error (about 2x better). Figure~\ref{fig:pose_est} shows some qualitative results where two viewpoints are provided in each example. The 3D bounding boxes computed using our poses are well aligned with the ground truth. Our method also accurately segments unarticulated regions compared to the GT mask, providing reliable correspondences for the subsequent PnP step in our method. The ablation study in the last two rows in Table~\ref{tab:pose_result} validates the effectiveness of our unarticulated region segmentation.

\subsection{Evaluation on interior recovery}
\label{subsec:eval_int_recovery}
To our knowledge, no prior works exist that perform interior recovery or articulated shape reconstruction from multi-articulation images. As such, we use the following baseline for comparison. 
Depth maps play a critical role in projecting plane masks to 3D space. GT depth maps can recover all interior planes when subject to multi-plane RANSAC plane fitting algorithm (and that's how ground truth 2D plane masks were obtained for training plane detection network). We extend this idea on the predicted depth maps and fit planes over them, per-view basis, and compare against the ground truth. Parameters for the RANSAC plane estimation algorithm are: threshold distance between planes is set to 5$e^{-3}$, and $\#$iters at 2K.

Table \ref{tab:baselines} presents quantitative results of this comparison over three \emph{randomly} chosen articulation view images per model and iterated hundred times to account for diverse view combinations. \rz{Our method outperforms the baseline on all three evaluation metrics.}
%
%
In particular, the geometry of the depth-associated 3D points does not allow for a meaningful plane estimation as seen via the normal angle difference and the mean of plane parameters difference.
Since our method is explicitly trained to detect planes in 2D and their associated 3D normals,
a better alignment to GT interior planes is \rz{expected}. 


\begin{figure}
\centering
\includegraphics[width=0.99\linewidth]{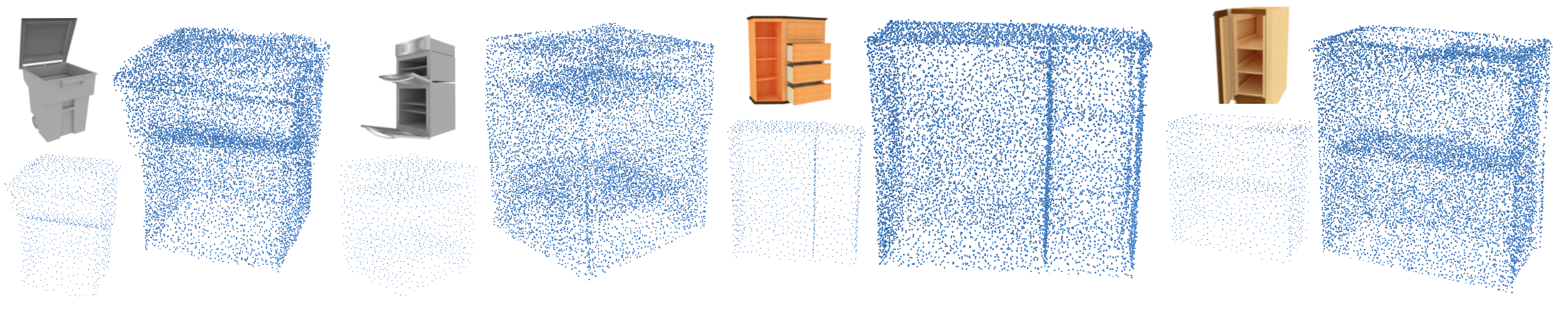}
\caption{Hallucinating interior geometry via shape completion. \rz{For each result, we show one reference articulation image and point clouds of both input (i.e., exterior shell) and output for shape completion.}}
\label{fig:shape_comp_results}
\end{figure}

\vspace{-5pt}

\rz{
\paragraph{Comparison to shape completion}
Finally, we investigate the hypothesis of using neural shape completion as an alternative to RoSI, with the important note that none of the current networks take image guidance as input. We chose the SOTA transformer-based shape completion network by \cite{zhou2022seedformer} and train it on our dataset to allow for {\em hallucinating\/} interior points from exterior shell point clouds. Figure \ref{fig:shape_comp_results} shows qualitative results on examples where we could observe completed interiors. However, hallucination only from the exteriors clearly does not work well as much of the interior structures are still missing from the outputs; see also a comparison on one-sided CD (measured {\em from} GT planes since shape completion results do not distinguish between interior and exterior points) in Table \ref{tab:baselines}. While such a direction comparison may not be entirely fair, it serves to reinforce the importance of image guidance which characterizes our new problem setting for RoSI.
}


\vspace{-5pt}

\begin{figure}
\centering
\includegraphics[width=0.99\linewidth]{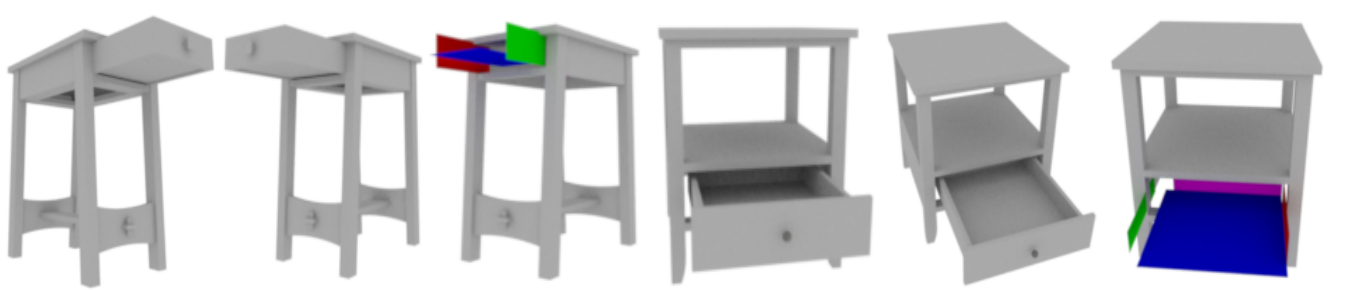}
\caption{RoSI results on untrained object category, table -- two input images and corresponding recovered interior planes}
\label{fig:untrained_generalization}
\end{figure}

\paragraph{Testing on untrained object category}
We assess the generalization capability of our method by testing on models unused during training, specifically, the Table category.
Figure \ref{fig:untrained_generalization} shows two such examples with their input images. Not only can RoSI recover interior planes, but it can also properly reproduce part motions in 3D.

\section{Conclusion, limitation, and future work}
\label{sec:future}

Our ultimate goal is to endow all existing 3D models with interior structures to truly complete them. 
To this end, we present RoSI, a learning framework to recover 3D shape interiors from multiple, sparse views of object articulation images. To the best of our knowledge, this is a new problem, while our current solution only represents a preliminary attempt which is still limited in several ways.

First, we can only recover, in 3D, those interior planes that are visible in at least one of the input images. RoSI currently lacks the ability to learn to reconstruct non-observable interior structures. 
Simple heuristics resorting to plane extension or symmetry-based structure completion may be employed. On the other hand, with training 3D shapes possessing full interiors, it is possible to re-formulate the learning problem to allow hallucinating unseen interior regions from their exterior contexts or other priors.

Second, our articulation/plane detection is performed in a per-view fashion 
and we use the predicted camera poses to project the inferred planes into a common frame. In doing so, no view consistency is enforced in learning the planes in 3D. While our design choice is motivated in part by the lack of image overlaps due to view sparsity, view consistency, if learned in the plane normal estimation step, could remove the dependency on a separate pose estimation network.

\begin{figure}
\centering
\includegraphics[width=0.99\linewidth]{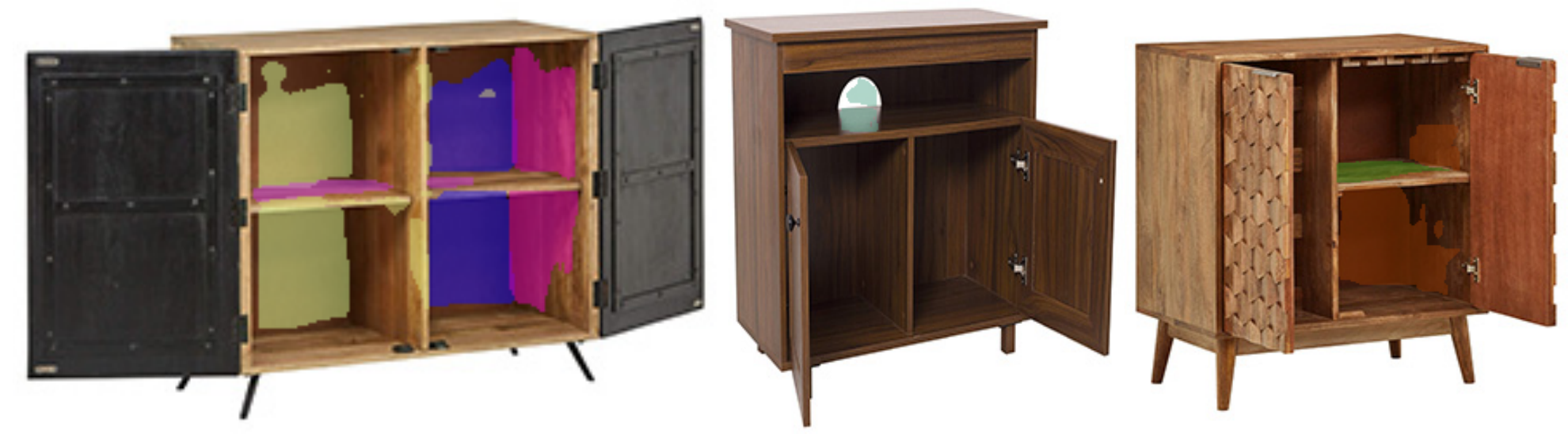}
\caption{Plane detection results on real images from the ABO dataset.}
\label{fig:abo_images}
\end{figure}

Third, we discovered a clear domain gap when testing on real articulation images, from the ABO dataset~\cite{collins2022abo}. Our method has been trained with synthetic data from PartNet-Mobility which lacks the realism and diversity in terms of textures, lighting, material, etc. Our pre-trained plane detection model fails to robustly detect most planes on these real images; see Figure \ref{fig:abo_images}. Such a domain gap could be addressed via data augmentation. 

Last but not least, \rz{while there is an abundance of online images showing part articulations,}
the dependency on having a quality 3D mesh shell during testing may not be practical. Foregoing this dependency would open up opportunities to develop neural architectures that can simultaneously \emph{reconstruct} the exterior surface and \emph{generate} interior structures from RGB images, particularly with plane representation, giving us means to create \emph{diverse} and \emph{functional} 3D assets.
\rz{In addition, most articulation images of online products come with accessories 
which can compromise our current plane detection scheme. This may be remedied with data augmentation.}

For future work, besides addressing the limitations discussed above, we may also analyze inter-plane, and more generally, structural relations between object parts, across views, as a means to improve both plane inference and integration; designing a learning scheme for such relations is a novel problem in itself. At last, generalizing our entire pipeline to accommodate more general 3D primitives and object types, are also necessary to fulfill our ultimate goal.

\bibliographystyle{ACM-Reference-Format}
\bibliography{10_references}


\begin{thebibliography}{48}


\ifx \showCODEN    \undefined \def \showCODEN     #1{\unskip}     \fi
\ifx \showDOI      \undefined \def \showDOI       #1{#1}\fi
\ifx \showISBNx    \undefined \def \showISBNx     #1{\unskip}     \fi
\ifx \showISBNxiii \undefined \def \showISBNxiii  #1{\unskip}     \fi
\ifx \showISSN     \undefined \def \showISSN      #1{\unskip}     \fi
\ifx \showLCCN     \undefined \def \showLCCN      #1{\unskip}     \fi
\ifx \shownote     \undefined \def \shownote      #1{#1}          \fi
\ifx \showarticletitle \undefined \def \showarticletitle #1{#1}   \fi
\ifx \showURL      \undefined \def \showURL       {\relax}        \fi
\providecommand\bibfield[2]{#2}
\providecommand\bibinfo[2]{#2}
\providecommand\natexlab[1]{#1}
\providecommand\showeprint[2][]{arXiv:#2}

\bibitem[\protect\citeauthoryear{Berger, Tagliasacchi, Seversky, Alliez,
  Guennebaud, Levine, Sharf, and Silva}{Berger et~al\mbox{.}}{2017}]%
        {berger2017survey}
\bibfield{author}{\bibinfo{person}{Matthew Berger}, \bibinfo{person}{Andrea
  Tagliasacchi}, \bibinfo{person}{Lee~M. Seversky}, \bibinfo{person}{Pierre
  Alliez}, \bibinfo{person}{Gael Guennebaud}, \bibinfo{person}{Joshua~A.
  Levine}, \bibinfo{person}{Andrei Sharf}, {and} \bibinfo{person}{Claudio~T.
  Silva}.} \bibinfo{year}{2017}\natexlab{}.
\newblock \showarticletitle{A Survey of Surface Reconstruction from Point
  Clouds}.
\newblock \bibinfo{journal}{\emph{Comput. Graph. Forum}}  \bibinfo{volume}{36}
  (\bibinfo{year}{2017}), \bibinfo{pages}{301--–329}.
\newblock
Issue 1.


\bibitem[\protect\citeauthoryear{Chang, Funkhouser, Guibas, Hanrahan, Huang,
  Li, Savarese, Savva, Song, Su, Xiao, Yi, and Yu}{Chang et~al\mbox{.}}{2015}]%
        {ShapeNet}
\bibfield{author}{\bibinfo{person}{Angel~X. Chang}, \bibinfo{person}{Thomas~A.
  Funkhouser}, \bibinfo{person}{Leonidas~J. Guibas}, \bibinfo{person}{Pat
  Hanrahan}, \bibinfo{person}{Qi{-}Xing Huang}, \bibinfo{person}{Zimo Li},
  \bibinfo{person}{Silvio Savarese}, \bibinfo{person}{Manolis Savva},
  \bibinfo{person}{Shuran Song}, \bibinfo{person}{Hao Su},
  \bibinfo{person}{Jianxiong Xiao}, \bibinfo{person}{Li Yi}, {and}
  \bibinfo{person}{Fisher Yu}.} \bibinfo{year}{2015}\natexlab{}.
\newblock \showarticletitle{ShapeNet: An Information-Rich 3D Model Repository}.
\newblock \bibinfo{journal}{\emph{CoRR}}  \bibinfo{volume}{abs/1512.03012}
  (\bibinfo{year}{2015}).
\newblock
\urldef\tempurl%
\url{http://arxiv.org/abs/1512.03012}
\showURL{%
\tempurl}


\bibitem[\protect\citeauthoryear{Collins, Goel, Deng, Luthra, Xu, Gundogdu,
  Zhang, Yago~Vicente, Dideriksen, Arora, Guillaumin, and Malik}{Collins
  et~al\mbox{.}}{2022}]%
        {collins2022abo}
\bibfield{author}{\bibinfo{person}{Jasmine Collins}, \bibinfo{person}{Shubham
  Goel}, \bibinfo{person}{Kenan Deng}, \bibinfo{person}{Achleshwar Luthra},
  \bibinfo{person}{Leon Xu}, \bibinfo{person}{Erhan Gundogdu},
  \bibinfo{person}{Xi Zhang}, \bibinfo{person}{Tomas~F Yago~Vicente},
  \bibinfo{person}{Thomas Dideriksen}, \bibinfo{person}{Himanshu Arora},
  \bibinfo{person}{Matthieu Guillaumin}, {and} \bibinfo{person}{Jitendra
  Malik}.} \bibinfo{year}{2022}\natexlab{}.
\newblock \showarticletitle{{ABO}: Dataset and Benchmarks for Real-World 3D
  Object Understanding}. In \bibinfo{booktitle}{\emph{CVPR}}.
\newblock


\bibitem[\protect\citeauthoryear{Desingh, Lu, Opipari, and Jenkins}{Desingh
  et~al\mbox{.}}{2019}]%
        {desingh2019factored}
\bibfield{author}{\bibinfo{person}{Karthik Desingh}, \bibinfo{person}{Shiyang
  Lu}, \bibinfo{person}{Anthony Opipari}, {and}
  \bibinfo{person}{Odest~Chadwicke Jenkins}.} \bibinfo{year}{2019}\natexlab{}.
\newblock \showarticletitle{Factored pose estimation of articulated objects
  using efficient nonparametric belief propagation}. In
  \bibinfo{booktitle}{\emph{International Conference on Robotics and
  Automation}}. \bibinfo{pages}{7221--7227}.
\newblock


\bibitem[\protect\citeauthoryear{Fischler and Bolles}{Fischler and
  Bolles}{1981}]%
        {fischler1981random}
\bibfield{author}{\bibinfo{person}{Martin~A Fischler} {and}
  \bibinfo{person}{Robert~C Bolles}.} \bibinfo{year}{1981}\natexlab{}.
\newblock \showarticletitle{Random sample consensus: a paradigm for model
  fitting with applications to image analysis and automated cartography}.
\newblock \bibinfo{journal}{\emph{Commun. ACM}} \bibinfo{volume}{24},
  \bibinfo{number}{6} (\bibinfo{year}{1981}), \bibinfo{pages}{381--395}.
\newblock


\bibitem[\protect\citeauthoryear{Fu, Jia, Gao, Gong, Zhao, Maybank, and Tao}{Fu
  et~al\mbox{.}}{2020}]%
        {fu20203dfuture}
\bibfield{author}{\bibinfo{person}{Huan Fu}, \bibinfo{person}{Rongfei Jia},
  \bibinfo{person}{Lin Gao}, \bibinfo{person}{Mingming Gong},
  \bibinfo{person}{Binqiang Zhao}, \bibinfo{person}{Steve Maybank}, {and}
  \bibinfo{person}{Dacheng Tao}.} \bibinfo{year}{2020}\natexlab{}.
\newblock \showarticletitle{3D-FUTURE: 3D Furniture shape with TextURE}.
\newblock \bibinfo{journal}{\emph{arXiv preprint arXiv:2009.09633}}
  (\bibinfo{year}{2020}).
\newblock


\bibitem[\protect\citeauthoryear{Furukawa, Curless, Seitz, and
  Szeliski}{Furukawa et~al\mbox{.}}{2009}]%
        {furukawa2009manhattan}
\bibfield{author}{\bibinfo{person}{Yasutaka Furukawa}, \bibinfo{person}{Brian
  Curless}, \bibinfo{person}{Steven~M Seitz}, {and} \bibinfo{person}{Richard
  Szeliski}.} \bibinfo{year}{2009}\natexlab{}.
\newblock \showarticletitle{Manhattan-world stereo}. In
  \bibinfo{booktitle}{\emph{CVPR}}. \bibinfo{pages}{1422--1429}.
\newblock


\bibitem[\protect\citeauthoryear{Gallup, Frahm, and Pollefeys}{Gallup
  et~al\mbox{.}}{2010}]%
        {gallup2010piecewise}
\bibfield{author}{\bibinfo{person}{David Gallup}, \bibinfo{person}{Jan-Michael
  Frahm}, {and} \bibinfo{person}{Marc Pollefeys}.}
  \bibinfo{year}{2010}\natexlab{}.
\newblock \showarticletitle{Piecewise planar and non-planar stereo for urban
  scene reconstruction}. In \bibinfo{booktitle}{\emph{CVPR}}.
  \bibinfo{pages}{1418--1425}.
\newblock


\bibitem[\protect\citeauthoryear{Hausman, Niekum, Osentoski, and
  Sukhatme}{Hausman et~al\mbox{.}}{2015}]%
        {hausman2015active}
\bibfield{author}{\bibinfo{person}{Karol Hausman}, \bibinfo{person}{Scott
  Niekum}, \bibinfo{person}{Sarah Osentoski}, {and} \bibinfo{person}{Gaurav~S
  Sukhatme}.} \bibinfo{year}{2015}\natexlab{}.
\newblock \showarticletitle{Active articulation model estimation through
  interactive perception}. In \bibinfo{booktitle}{\emph{International
  Conference on Robotics and Automation}}. \bibinfo{pages}{3305--3312}.
\newblock


\bibitem[\protect\citeauthoryear{He, Gkioxari, Doll{\'a}r, and Girshick}{He
  et~al\mbox{.}}{2017}]%
        {he2017mask}
\bibfield{author}{\bibinfo{person}{Kaiming He}, \bibinfo{person}{Georgia
  Gkioxari}, \bibinfo{person}{Piotr Doll{\'a}r}, {and} \bibinfo{person}{Ross
  Girshick}.} \bibinfo{year}{2017}\natexlab{}.
\newblock \showarticletitle{Mask {R-CNN}}. In \bibinfo{booktitle}{\emph{ICCV}}.
  \bibinfo{pages}{2961--2969}.
\newblock


\bibitem[\protect\citeauthoryear{Huang, Yu, Xu, Ni, and Le}{Huang
  et~al\mbox{.}}{2020}]%
        {huang2020pf}
\bibfield{author}{\bibinfo{person}{Zitian Huang}, \bibinfo{person}{Yikuan Yu},
  \bibinfo{person}{Jiawen Xu}, \bibinfo{person}{Feng Ni}, {and}
  \bibinfo{person}{Xinyi Le}.} \bibinfo{year}{2020}\natexlab{}.
\newblock \showarticletitle{Pf-net: Point fractal network for 3d point cloud
  completion}. In \bibinfo{booktitle}{\emph{Proceedings of the IEEE/CVF
  conference on computer vision and pattern recognition}}.
  \bibinfo{pages}{7662--7670}.
\newblock


\bibitem[\protect\citeauthoryear{Jakob, Speierer, Roussel, Nimier-David,
  Vicini, Zeltner, Nicolet, Crespo, Leroy, and Zhang}{Jakob
  et~al\mbox{.}}{2022}]%
        {jakob2022mitsuba3}
\bibfield{author}{\bibinfo{person}{Wenzel Jakob}, \bibinfo{person}{Sébastien
  Speierer}, \bibinfo{person}{Nicolas Roussel}, \bibinfo{person}{Merlin
  Nimier-David}, \bibinfo{person}{Delio Vicini}, \bibinfo{person}{Tizian
  Zeltner}, \bibinfo{person}{Baptiste Nicolet}, \bibinfo{person}{Miguel
  Crespo}, \bibinfo{person}{Vincent Leroy}, {and} \bibinfo{person}{Ziyi
  Zhang}.} \bibinfo{year}{2022}\natexlab{}.
\newblock \bibinfo{booktitle}{\emph{Mitsuba 3 renderer}}.
\newblock
\newblock
\shownote{https://mitsuba-renderer.org.}


\bibitem[\protect\citeauthoryear{Jiang, Mao, Savva, and Chang}{Jiang
  et~al\mbox{.}}{2022}]%
        {jiang2022opd}
\bibfield{author}{\bibinfo{person}{Hanxiao Jiang}, \bibinfo{person}{Yongsen
  Mao}, \bibinfo{person}{Manolis Savva}, {and} \bibinfo{person}{Angel~X
  Chang}.} \bibinfo{year}{2022}\natexlab{}.
\newblock \showarticletitle{OPD: Single-view 3D Openable Part Detection}.
\newblock  (\bibinfo{year}{2022}).
\newblock


\bibitem[\protect\citeauthoryear{Jiang, Liu, Schulter, Wang, and
  Chandraker}{Jiang et~al\mbox{.}}{2020}]%
        {jiang2020peek}
\bibfield{author}{\bibinfo{person}{Ziyu Jiang}, \bibinfo{person}{Buyu Liu},
  \bibinfo{person}{Samuel Schulter}, \bibinfo{person}{Zhangyang Wang}, {and}
  \bibinfo{person}{Manmohan Chandraker}.} \bibinfo{year}{2020}\natexlab{}.
\newblock \showarticletitle{Peek-a-boo: Occlusion reasoning in indoor scenes
  with plane representations}. In \bibinfo{booktitle}{\emph{CVPR}}.
  \bibinfo{pages}{113--121}.
\newblock


\bibitem[\protect\citeauthoryear{Jin, Qian, Owens, and Fouhey}{Jin
  et~al\mbox{.}}{2021}]%
        {jin2021planar}
\bibfield{author}{\bibinfo{person}{Linyi Jin}, \bibinfo{person}{Shengyi Qian},
  \bibinfo{person}{Andrew Owens}, {and} \bibinfo{person}{David~F Fouhey}.}
  \bibinfo{year}{2021}\natexlab{}.
\newblock \showarticletitle{Planar surface reconstruction from sparse views}.
  In \bibinfo{booktitle}{\emph{ICCV}}. \bibinfo{pages}{12991--13000}.
\newblock


\bibitem[\protect\citeauthoryear{Katz and Brock}{Katz and Brock}{2008}]%
        {katz2008manipulating}
\bibfield{author}{\bibinfo{person}{Dov Katz} {and} \bibinfo{person}{Oliver
  Brock}.} \bibinfo{year}{2008}\natexlab{}.
\newblock \showarticletitle{Manipulating articulated objects with interactive
  perception}. In \bibinfo{booktitle}{\emph{International Conference on
  Robotics and Automation}}. \bibinfo{pages}{272--277}.
\newblock


\bibitem[\protect\citeauthoryear{Li, Wang, Yi, Guibas, Abbott, and Song}{Li
  et~al\mbox{.}}{2020}]%
        {li2020category}
\bibfield{author}{\bibinfo{person}{Xiaolong Li}, \bibinfo{person}{He Wang},
  \bibinfo{person}{Li Yi}, \bibinfo{person}{Leonidas~J Guibas},
  \bibinfo{person}{A~Lynn Abbott}, {and} \bibinfo{person}{Shuran Song}.}
  \bibinfo{year}{2020}\natexlab{}.
\newblock \showarticletitle{Category-level articulated object pose estimation}.
  In \bibinfo{booktitle}{\emph{CVPR}}. \bibinfo{pages}{3706--3715}.
\newblock


\bibitem[\protect\citeauthoryear{Liu, Kim, Gu, Furukawa, and Kautz}{Liu
  et~al\mbox{.}}{2019}]%
        {liu2019planercnn}
\bibfield{author}{\bibinfo{person}{Chen Liu}, \bibinfo{person}{Kihwan Kim},
  \bibinfo{person}{Jinwei Gu}, \bibinfo{person}{Yasutaka Furukawa}, {and}
  \bibinfo{person}{Jan Kautz}.} \bibinfo{year}{2019}\natexlab{}.
\newblock \showarticletitle{{PlaneRCNN: 3D} plane detection and reconstruction
  from a single image}. In \bibinfo{booktitle}{\emph{CVPR}}.
  \bibinfo{pages}{4450--4459}.
\newblock


\bibitem[\protect\citeauthoryear{Liu, Yang, Ceylan, Yumer, and Furukawa}{Liu
  et~al\mbox{.}}{2018}]%
        {liu2018planenet}
\bibfield{author}{\bibinfo{person}{Chen Liu}, \bibinfo{person}{Jimei Yang},
  \bibinfo{person}{Duygu Ceylan}, \bibinfo{person}{Ersin Yumer}, {and}
  \bibinfo{person}{Yasutaka Furukawa}.} \bibinfo{year}{2018}\natexlab{}.
\newblock \showarticletitle{Planenet: Piece-wise planar reconstruction from a
  single rgb image}. In \bibinfo{booktitle}{\emph{CVPR}}.
  \bibinfo{pages}{2579--2588}.
\newblock


\bibitem[\protect\citeauthoryear{Liu, Xue, Xu, Fu, and Lu}{Liu
  et~al\mbox{.}}{2022}]%
        {liu2022toward}
\bibfield{author}{\bibinfo{person}{Liu Liu}, \bibinfo{person}{Han Xue},
  \bibinfo{person}{Wenqiang Xu}, \bibinfo{person}{Haoyuan Fu}, {and}
  \bibinfo{person}{Cewu Lu}.} \bibinfo{year}{2022}\natexlab{}.
\newblock \showarticletitle{Toward Real-World Category-Level Articulation Pose
  Estimation}.
\newblock \bibinfo{journal}{\emph{IEEE Transactions on Image Processing}}
  \bibinfo{volume}{31} (\bibinfo{year}{2022}), \bibinfo{pages}{1072--1083}.
\newblock


\bibitem[\protect\citeauthoryear{Liu, Qiu, Wang, Hager, and Yuille}{Liu
  et~al\mbox{.}}{2020}]%
        {liu2020nothing}
\bibfield{author}{\bibinfo{person}{Qihao Liu}, \bibinfo{person}{Weichao Qiu},
  \bibinfo{person}{Weiyao Wang}, \bibinfo{person}{Gregory~D Hager}, {and}
  \bibinfo{person}{Alan~L Yuille}.} \bibinfo{year}{2020}\natexlab{}.
\newblock \showarticletitle{Nothing but geometric constraints: A model-free
  method for articulated object pose estimation}.
\newblock \bibinfo{journal}{\emph{arXiv preprint arXiv:2012.00088}}
  (\bibinfo{year}{2020}).
\newblock


\bibitem[\protect\citeauthoryear{Michel, Krull, Brachmann, Yang, Gumhold, and
  Rother}{Michel et~al\mbox{.}}{2015}]%
        {michel2015pose}
\bibfield{author}{\bibinfo{person}{Frank Michel}, \bibinfo{person}{Alexander
  Krull}, \bibinfo{person}{Eric Brachmann}, \bibinfo{person}{Michael~Ying
  Yang}, \bibinfo{person}{Stefan Gumhold}, {and} \bibinfo{person}{Carsten
  Rother}.} \bibinfo{year}{2015}\natexlab{}.
\newblock \showarticletitle{Pose Estimation of Kinematic Chain Instances via
  Object Coordinate Regression.}. In \bibinfo{booktitle}{\emph{BMVC}}.
  \bibinfo{pages}{181--1}.
\newblock


\bibitem[\protect\citeauthoryear{Mo, Zhu, Chang, Yi, Tripathi, Guibas, and
  Su}{Mo et~al\mbox{.}}{2019}]%
        {mo2019partnet}
\bibfield{author}{\bibinfo{person}{Kaichun Mo}, \bibinfo{person}{Shilin Zhu},
  \bibinfo{person}{Angel~X. Chang}, \bibinfo{person}{Li Yi},
  \bibinfo{person}{Subarna Tripathi}, \bibinfo{person}{Leonidas~J. Guibas},
  {and} \bibinfo{person}{Hao Su}.} \bibinfo{year}{2019}\natexlab{}.
\newblock \showarticletitle{{PartNet}: A Large-Scale Benchmark for Fine-Grained
  and Hierarchical Part-Level {3D} Object Understanding}. In
  \bibinfo{booktitle}{\emph{CVPR}}.
\newblock


\bibitem[\protect\citeauthoryear{Mu, Qiu, Kortylewski, Yuille, Vasconcelos, and
  Wang}{Mu et~al\mbox{.}}{2021}]%
        {mu2021sdf}
\bibfield{author}{\bibinfo{person}{Jiteng Mu}, \bibinfo{person}{Weichao Qiu},
  \bibinfo{person}{Adam Kortylewski}, \bibinfo{person}{Alan Yuille},
  \bibinfo{person}{Nuno Vasconcelos}, {and} \bibinfo{person}{Xiaolong Wang}.}
  \bibinfo{year}{2021}\natexlab{}.
\newblock \showarticletitle{A-sdf: Learning disentangled signed distance
  functions for articulated shape representation}. In
  \bibinfo{booktitle}{\emph{ICCV}}. \bibinfo{pages}{13001--13011}.
\newblock


\bibitem[\protect\citeauthoryear{Pan, Chen, Cai, Zhang, Zhao, Yi, and Liu}{Pan
  et~al\mbox{.}}{2021}]%
        {pan2021variational}
\bibfield{author}{\bibinfo{person}{Liang Pan}, \bibinfo{person}{Xinyi Chen},
  \bibinfo{person}{Zhongang Cai}, \bibinfo{person}{Junzhe Zhang},
  \bibinfo{person}{Haiyu Zhao}, \bibinfo{person}{Shuai Yi}, {and}
  \bibinfo{person}{Ziwei Liu}.} \bibinfo{year}{2021}\natexlab{}.
\newblock \showarticletitle{Variational relational point completion network}.
  In \bibinfo{booktitle}{\emph{Proceedings of the IEEE/CVF conference on
  computer vision and pattern recognition}}. \bibinfo{pages}{8524--8533}.
\newblock


\bibitem[\protect\citeauthoryear{Pavlasek, Lewis, Desingh, and
  Jenkins}{Pavlasek et~al\mbox{.}}{2020}]%
        {pavlasek2020parts}
\bibfield{author}{\bibinfo{person}{Jana Pavlasek}, \bibinfo{person}{Stanley
  Lewis}, \bibinfo{person}{Karthik Desingh}, {and}
  \bibinfo{person}{Odest~Chadwicke Jenkins}.} \bibinfo{year}{2020}\natexlab{}.
\newblock \showarticletitle{Parts-based articulated object localization in
  clutter using belief propagation}. In \bibinfo{booktitle}{\emph{2020 IEEE/RSJ
  International Conference on Intelligent Robots and Systems (IROS)}}.
  \bibinfo{pages}{10595--10602}.
\newblock


\bibitem[\protect\citeauthoryear{Qian, Jin, Rockwell, Chen, and Fouhey}{Qian
  et~al\mbox{.}}{2022}]%
        {qian2022understanding}
\bibfield{author}{\bibinfo{person}{Shengyi Qian}, \bibinfo{person}{Linyi Jin},
  \bibinfo{person}{Chris Rockwell}, \bibinfo{person}{Siyi Chen}, {and}
  \bibinfo{person}{David~F Fouhey}.} \bibinfo{year}{2022}\natexlab{}.
\newblock \showarticletitle{Understanding 3D Object Articulation in Internet
  Videos}. In \bibinfo{booktitle}{\emph{CVPR}}. \bibinfo{pages}{1599--1609}.
\newblock


\bibitem[\protect\citeauthoryear{Ranftl, Bochkovskiy, and Koltun}{Ranftl
  et~al\mbox{.}}{2021}]%
        {ranftl2021vision}
\bibfield{author}{\bibinfo{person}{Ren{\'e} Ranftl}, \bibinfo{person}{Alexey
  Bochkovskiy}, {and} \bibinfo{person}{Vladlen Koltun}.}
  \bibinfo{year}{2021}\natexlab{}.
\newblock \showarticletitle{Vision transformers for dense prediction}. In
  \bibinfo{booktitle}{\emph{ICCV}}. \bibinfo{pages}{12179--12188}.
\newblock


\bibitem[\protect\citeauthoryear{Ronneberger, Fischer, and Brox}{Ronneberger
  et~al\mbox{.}}{2015}]%
        {ronneberger2015u}
\bibfield{author}{\bibinfo{person}{Olaf Ronneberger}, \bibinfo{person}{Philipp
  Fischer}, {and} \bibinfo{person}{Thomas Brox}.}
  \bibinfo{year}{2015}\natexlab{}.
\newblock \showarticletitle{U-net: Convolutional networks for biomedical image
  segmentation}. In \bibinfo{booktitle}{\emph{International Conference on
  Medical image computing and computer-assisted intervention}}. Springer,
  \bibinfo{pages}{234--241}.
\newblock


\bibitem[\protect\citeauthoryear{Silberman, Hoiem, Kohli, and Fergus}{Silberman
  et~al\mbox{.}}{2012}]%
        {silberman2012indoor}
\bibfield{author}{\bibinfo{person}{Nathan Silberman}, \bibinfo{person}{Derek
  Hoiem}, \bibinfo{person}{Pushmeet Kohli}, {and} \bibinfo{person}{Rob
  Fergus}.} \bibinfo{year}{2012}\natexlab{}.
\newblock \showarticletitle{Indoor segmentation and support inference from rgbd
  images}. In \bibinfo{booktitle}{\emph{ECCV}}. Springer,
  \bibinfo{pages}{746--760}.
\newblock


\bibitem[\protect\citeauthoryear{Sinha, Steedly, and Szeliski}{Sinha
  et~al\mbox{.}}{2009}]%
        {sinha2009piecewise}
\bibfield{author}{\bibinfo{person}{Sudipta Sinha}, \bibinfo{person}{Drew
  Steedly}, {and} \bibinfo{person}{Rick Szeliski}.}
  \bibinfo{year}{2009}\natexlab{}.
\newblock \showarticletitle{Piecewise planar stereo for image-based rendering}.
  In \bibinfo{booktitle}{\emph{ICCV}}. \bibinfo{pages}{1881--1888}.
\newblock


\bibitem[\protect\citeauthoryear{Tan, Xue, Bai, Wu, and Xia}{Tan
  et~al\mbox{.}}{2021}]%
        {tan2021planetr}
\bibfield{author}{\bibinfo{person}{Bin Tan}, \bibinfo{person}{Nan Xue},
  \bibinfo{person}{Song Bai}, \bibinfo{person}{Tianfu Wu}, {and}
  \bibinfo{person}{Gui-Song Xia}.} \bibinfo{year}{2021}\natexlab{}.
\newblock \showarticletitle{Planetr: Structure-guided transformers for 3d plane
  recovery}. In \bibinfo{booktitle}{\emph{ICCV}}. \bibinfo{pages}{4186--4195}.
\newblock


\bibitem[\protect\citeauthoryear{Wang and Solomon}{Wang and Solomon}{2019}]%
        {wang2019deep}
\bibfield{author}{\bibinfo{person}{Yue Wang} {and} \bibinfo{person}{Justin~M
  Solomon}.} \bibinfo{year}{2019}\natexlab{}.
\newblock \showarticletitle{Deep closest point: Learning representations for
  point cloud registration}. In \bibinfo{booktitle}{\emph{CVPR}}.
  \bibinfo{pages}{3523--3532}.
\newblock


\bibitem[\protect\citeauthoryear{Wang, Sun, Liu, Sarma, Bronstein, and
  Solomon}{Wang et~al\mbox{.}}{2019}]%
        {wang2019dynamic}
\bibfield{author}{\bibinfo{person}{Yue Wang}, \bibinfo{person}{Yongbin Sun},
  \bibinfo{person}{Ziwei Liu}, \bibinfo{person}{Sanjay~E Sarma},
  \bibinfo{person}{Michael~M Bronstein}, {and} \bibinfo{person}{Justin~M
  Solomon}.} \bibinfo{year}{2019}\natexlab{}.
\newblock \showarticletitle{Dynamic graph {CNN} for learning on point clouds}.
\newblock \bibinfo{journal}{\emph{ACM TOG}} \bibinfo{volume}{38},
  \bibinfo{number}{5} (\bibinfo{year}{2019}), \bibinfo{pages}{1--12}.
\newblock


\bibitem[\protect\citeauthoryear{Wei, Chabra, Ma, Lassner, Zollh{\"o}fer,
  Rusinkiewicz, Sweeney, Newcombe, and Slavcheva}{Wei et~al\mbox{.}}{2022}]%
        {wei2022self}
\bibfield{author}{\bibinfo{person}{Fangyin Wei}, \bibinfo{person}{Rohan
  Chabra}, \bibinfo{person}{Lingni Ma}, \bibinfo{person}{Christoph Lassner},
  \bibinfo{person}{Michael Zollh{\"o}fer}, \bibinfo{person}{Szymon
  Rusinkiewicz}, \bibinfo{person}{Chris Sweeney}, \bibinfo{person}{Richard
  Newcombe}, {and} \bibinfo{person}{Mira Slavcheva}.}
  \bibinfo{year}{2022}\natexlab{}.
\newblock \showarticletitle{Self-supervised Neural Articulated Shape and
  Appearance Models}. In \bibinfo{booktitle}{\emph{CVPR}}.
  \bibinfo{pages}{15816--15826}.
\newblock


\bibitem[\protect\citeauthoryear{Weng, Wang, Zhou, Qin, Duan, Fan, Chen, Su,
  and Guibas}{Weng et~al\mbox{.}}{2021}]%
        {weng2021captra}
\bibfield{author}{\bibinfo{person}{Yijia Weng}, \bibinfo{person}{He Wang},
  \bibinfo{person}{Qiang Zhou}, \bibinfo{person}{Yuzhe Qin},
  \bibinfo{person}{Yueqi Duan}, \bibinfo{person}{Qingnan Fan},
  \bibinfo{person}{Baoquan Chen}, \bibinfo{person}{Hao Su}, {and}
  \bibinfo{person}{Leonidas~J Guibas}.} \bibinfo{year}{2021}\natexlab{}.
\newblock \showarticletitle{Captra: Category-level pose tracking for rigid and
  articulated objects from point clouds}. In \bibinfo{booktitle}{\emph{ICCV}}.
  \bibinfo{pages}{13209--13218}.
\newblock


\bibitem[\protect\citeauthoryear{Xiang, Qin, Mo, Xia, Zhu, Liu, Liu, Jiang,
  Yuan, Wang, et~al\mbox{.}}{Xiang et~al\mbox{.}}{2020}]%
        {xiang2020sapien}
\bibfield{author}{\bibinfo{person}{Fanbo Xiang}, \bibinfo{person}{Yuzhe Qin},
  \bibinfo{person}{Kaichun Mo}, \bibinfo{person}{Yikuan Xia},
  \bibinfo{person}{Hao Zhu}, \bibinfo{person}{Fangchen Liu},
  \bibinfo{person}{Minghua Liu}, \bibinfo{person}{Hanxiao Jiang},
  \bibinfo{person}{Yifu Yuan}, \bibinfo{person}{He Wang}, {et~al\mbox{.}}}
  \bibinfo{year}{2020}\natexlab{}.
\newblock \showarticletitle{Sapien: A simulated part-based interactive
  environment}. In \bibinfo{booktitle}{\emph{CVPR}}.
  \bibinfo{pages}{11097--11107}.
\newblock


\bibitem[\protect\citeauthoryear{Xiang, Wen, Liu, Cao, Wan, Zheng, and
  Han}{Xiang et~al\mbox{.}}{2022}]%
        {xiang2022snowflake}
\bibfield{author}{\bibinfo{person}{Peng Xiang}, \bibinfo{person}{Xin Wen},
  \bibinfo{person}{Yu-Shen Liu}, \bibinfo{person}{Yan-Pei Cao},
  \bibinfo{person}{Pengfei Wan}, \bibinfo{person}{Wen Zheng}, {and}
  \bibinfo{person}{Zhizhong Han}.} \bibinfo{year}{2022}\natexlab{}.
\newblock \showarticletitle{Snowflake Point Deconvolution for Point Cloud
  Completion and Generation with Skip-Transformer}.
\newblock \bibinfo{journal}{\emph{arXiv preprint arXiv:2202.09367}}
  (\bibinfo{year}{2022}).
\newblock


\bibitem[\protect\citeauthoryear{Xiao, Qiu, Langlois, Aubry, and Marlet}{Xiao
  et~al\mbox{.}}{2019}]%
        {Xiao2019PoseFromShape}
\bibfield{author}{\bibinfo{person}{Yang Xiao}, \bibinfo{person}{Xuchong Qiu},
  \bibinfo{person}{Pierre{-}Alain Langlois}, \bibinfo{person}{Mathieu Aubry},
  {and} \bibinfo{person}{Renaud Marlet}.} \bibinfo{year}{2019}\natexlab{}.
\newblock \showarticletitle{Pose from Shape: Deep Pose Estimation for Arbitrary
  {3D} Objects}. In \bibinfo{booktitle}{\emph{British Machine Vision
  Conference}}.
\newblock


\bibitem[\protect\citeauthoryear{Xue, Liu, Xu, Fu, and Lu}{Xue
  et~al\mbox{.}}{2021}]%
        {XueLXFL21OMAD}
\bibfield{author}{\bibinfo{person}{Han Xue}, \bibinfo{person}{Liu Liu},
  \bibinfo{person}{Wenqiang Xu}, \bibinfo{person}{Haoyuan Fu}, {and}
  \bibinfo{person}{Cewu Lu}.} \bibinfo{year}{2021}\natexlab{}.
\newblock \showarticletitle{{OMAD:} Object Model with Articulated Deformations
  for Pose Estimation and Retrieval}. In \bibinfo{booktitle}{\emph{32nd British
  Machine Vision Conference 2021}}. \bibinfo{publisher}{BMVA Press},
  \bibinfo{pages}{366}.
\newblock


\bibitem[\protect\citeauthoryear{Yan, Sharf, Lin, Huang, and Chen}{Yan
  et~al\mbox{.}}{2014}]%
        {yan2014proactive}
\bibfield{author}{\bibinfo{person}{Feilong Yan}, \bibinfo{person}{Andrei
  Sharf}, \bibinfo{person}{Wenzhen Lin}, \bibinfo{person}{Hui Huang}, {and}
  \bibinfo{person}{Baoquan Chen}.} \bibinfo{year}{2014}\natexlab{}.
\newblock \showarticletitle{Proactive 3D scanning of inaccessible parts}.
\newblock \bibinfo{journal}{\emph{ACM TOG}} \bibinfo{volume}{33},
  \bibinfo{number}{4} (\bibinfo{year}{2014}), \bibinfo{pages}{157}.
\newblock


\bibitem[\protect\citeauthoryear{Yan, Lin, Mitra, Lischinski, Cohen-Or, and
  Huang}{Yan et~al\mbox{.}}{2022}]%
        {yan2022shapeformer}
\bibfield{author}{\bibinfo{person}{Xingguang Yan}, \bibinfo{person}{Liqiang
  Lin}, \bibinfo{person}{Niloy~J Mitra}, \bibinfo{person}{Dani Lischinski},
  \bibinfo{person}{Daniel Cohen-Or}, {and} \bibinfo{person}{Hui Huang}.}
  \bibinfo{year}{2022}\natexlab{}.
\newblock \showarticletitle{Shapeformer: Transformer-based shape completion via
  sparse representation}. In \bibinfo{booktitle}{\emph{Proceedings of the
  IEEE/CVF Conference on Computer Vision and Pattern Recognition}}.
  \bibinfo{pages}{6239--6249}.
\newblock


\bibitem[\protect\citeauthoryear{Yu, Rao, Wang, Liu, Lu, and Zhou}{Yu
  et~al\mbox{.}}{2021}]%
        {yu2021pointr}
\bibfield{author}{\bibinfo{person}{Xumin Yu}, \bibinfo{person}{Yongming Rao},
  \bibinfo{person}{Ziyi Wang}, \bibinfo{person}{Zuyan Liu},
  \bibinfo{person}{Jiwen Lu}, {and} \bibinfo{person}{Jie Zhou}.}
  \bibinfo{year}{2021}\natexlab{}.
\newblock \showarticletitle{Pointr: Diverse point cloud completion with
  geometry-aware transformers}. In \bibinfo{booktitle}{\emph{Proceedings of the
  IEEE/CVF international conference on computer vision}}.
  \bibinfo{pages}{12498--12507}.
\newblock


\bibitem[\protect\citeauthoryear{Yu, Zheng, Lian, Zhou, and Gao}{Yu
  et~al\mbox{.}}{2019}]%
        {yu2019single}
\bibfield{author}{\bibinfo{person}{Zehao Yu}, \bibinfo{person}{Jia Zheng},
  \bibinfo{person}{Dongze Lian}, \bibinfo{person}{Zihan Zhou}, {and}
  \bibinfo{person}{Shenghua Gao}.} \bibinfo{year}{2019}\natexlab{}.
\newblock \showarticletitle{Single-image piece-wise planar 3d reconstruction
  via associative embedding}. In \bibinfo{booktitle}{\emph{CVPR}}.
  \bibinfo{pages}{1029--1037}.
\newblock


\bibitem[\protect\citeauthoryear{Yuan, Khot, Held, Mertz, and Hebert}{Yuan
  et~al\mbox{.}}{2018}]%
        {yuan2018pcn}
\bibfield{author}{\bibinfo{person}{Wentao Yuan}, \bibinfo{person}{Tejas Khot},
  \bibinfo{person}{David Held}, \bibinfo{person}{Christoph Mertz}, {and}
  \bibinfo{person}{Martial Hebert}.} \bibinfo{year}{2018}\natexlab{}.
\newblock \showarticletitle{Pcn: Point completion network}. In
  \bibinfo{booktitle}{\emph{2018 International Conference on 3D Vision (3DV)}}.
  IEEE, \bibinfo{pages}{728--737}.
\newblock


\bibitem[\protect\citeauthoryear{Zebedin, Bauer, Karner, and Bischof}{Zebedin
  et~al\mbox{.}}{2008}]%
        {zebedin2008fusion}
\bibfield{author}{\bibinfo{person}{Lukas Zebedin}, \bibinfo{person}{Joachim
  Bauer}, \bibinfo{person}{Konrad Karner}, {and} \bibinfo{person}{Horst
  Bischof}.} \bibinfo{year}{2008}\natexlab{}.
\newblock \showarticletitle{Fusion of feature-and area-based information for
  urban buildings modeling from aerial imagery}. In
  \bibinfo{booktitle}{\emph{ECCV}}. Springer, \bibinfo{pages}{873--886}.
\newblock


\bibitem[\protect\citeauthoryear{Zhang, Litany, Sridhar, and Guibas}{Zhang
  et~al\mbox{.}}{2021}]%
        {zhang2021strobenet}
\bibfield{author}{\bibinfo{person}{Ge Zhang}, \bibinfo{person}{Or Litany},
  \bibinfo{person}{Srinath Sridhar}, {and} \bibinfo{person}{Leonidas Guibas}.}
  \bibinfo{year}{2021}\natexlab{}.
\newblock \showarticletitle{StrobeNet: Category-Level Multiview Reconstruction
  of Articulated Objects}.
\newblock \bibinfo{journal}{\emph{unpublished arXiv preprint arXiv:2105.08016}}
  (\bibinfo{year}{2021}).
\newblock


\bibitem[\protect\citeauthoryear{Zhou, Cao, Chu, Zhu, Lu, Tai, and Wang}{Zhou
  et~al\mbox{.}}{2022}]%
        {zhou2022seedformer}
\bibfield{author}{\bibinfo{person}{Haoran Zhou}, \bibinfo{person}{Yun Cao},
  \bibinfo{person}{Wenqing Chu}, \bibinfo{person}{Junwei Zhu},
  \bibinfo{person}{Tong Lu}, \bibinfo{person}{Ying Tai}, {and}
  \bibinfo{person}{Chengjie Wang}.} \bibinfo{year}{2022}\natexlab{}.
\newblock \showarticletitle{Seedformer: Patch seeds based point cloud
  completion with upsample transformer}. In \bibinfo{booktitle}{\emph{Computer
  Vision--ECCV 2022: 17th European Conference, Tel Aviv, Israel, October
  23--27, 2022, Proceedings, Part III}}. Springer, \bibinfo{pages}{416--432}.
\newblock


\end{thebibliography}

\end{document}